\documentclass[sigconf]{acmart}
\usepackage{booktabs}
\usepackage{multirow}
\usepackage{array}
\usepackage{makecell}
\usepackage{tabularx}
\usepackage{colortbl}
\usepackage{xcolor}
\usepackage{graphicx}
\usepackage{subcaption}
\usepackage{multirow}
\usepackage{booktabs}
\usepackage{makecell}
\usepackage{graphicx}
\usepackage{subcaption}
\usepackage{xr}
\usepackage{balance}
\newenvironment{lstlisting}{}{}
\AtBeginDocument{%
  }

\setcopyright{acmlicensed}
\copyrightyear{2026}
\acmYear{2026}
\setcopyright{cc}
\setcctype{by}
\acmConference[MM '26]{Proceedings of the 34th ACM International Conference on Multimedia}{November 10--14, 2026}{Rio de Janeiro, Brazil}
\acmBooktitle{Proceedings of the 34th ACM International Conference on Multimedia (MM '26), November 10--14, 2026, Rio de Janeiro, Brazil}
\acmDOI{10.1145/3767308.3836364}
\acmISBN{979-8-4007-2213-4/2026/11}

\begin{document}

\title{DDVR-Net: Degradation-Aware Decoupling and Visibility-Guided Restoration for Historical Document Recognition}


\author{Jiu Sha}
\email{shajiu@muc.edu.cn}
\orcid{0000-0002-3396-7520}
%\correspondingauthor
\affiliation{%
  \institution{Institute of National Security, Minzu University of China}
  \city{Beijing}
  \country{China}
}

\affiliation{%
  \institution{School of Information Engineering, Minzu University of China}
  \city{Beijing}
  \postcode{100081}
  \country{China}
}

\author{Yu Weng}
\email{wengyu@muc.edu.cn}
\correspondingauthor
\orcid{0000-0002-0787-550X}
%\correspondingauthor
\affiliation{%
  \institution{School of Information Engineering, Minzu University of China}
  \city{Beijing}
  \country{China}
}

\affiliation{%
  \institution{Institute of National Security, Minzu University of China}
  \city{Beijing}
  \country{China}
}

\author{Mengxiao Zhu}
\email{zhumx@ncut.edu.cn}
\correspondingauthor
\affiliation{%
  \institution{School of Artificial Intelligence and Computer Science, North China University of Technology}
  \city{Beijing}
  \country{China}
}

% \affiliation{%

% \affiliation{%

% \affiliation{%

% \affiliation{%

% \affiliation{%

% \correspondingauthor
% \affiliation{%

\renewcommand{\shortauthors}{Jiu Sha and Mengxiao Zhu.}

\begin{abstract}
Historical document analysis is crucial for cultural heritage preservation and low-resource language digitization. However, real historical documents often suffer from severe degradation, such as ink erosion and structural damage, causing mixed-visibility character sequences. Conventional OCR systems struggle in such evidence-limited scenarios, while contextual language models may over-correct visually clear characters. To address these challenges, we propose DDVR-Net, a degradation-aware decoupling and visibility-guided restoration framework for historical document recognition. DDVR-Net decouples intrinsic glyph structures from physical corruption at the feature level and predicts local visibility to estimate character reliability. A contextual decoder is selectively activated through a visibility-guided gate, restoring degraded characters while preserving clear visual recognition. Furthermore, we construct a low-resource multilingual benchmark covering Tibetan, Mongolian, and Uyghur scripts with 187000 document pairs, combining large-scale synthetic data and progressive training with real-world annotations to reduce the synthetic-to-real gap. Extensive experiments demonstrate that DDVR-Net consistently outperforms existing methods across three languages and diverse reading orders, providing an effective framework and benchmark for low-resource historical document recognition. \textbf{Code and Benchmark:} \url{https://github.com/Shajiu/DDVR-Net}
\end{abstract}



%, Degradation-Aware Learning, Visibility-Guided Restoration




\maketitle

% \begin{figure}[t]
%     \centering
%     \begin{subfigure}{0.153\textwidth}
%         \centering
%     \includegraphics[width=\linewidth, height=4cm]{figures/huihewen.jpg}
%         \caption{Uyghur}
%         \label{fig:sub_a}
%     \end{subfigure}
%     \hfill 
%     \begin{subfigure}{0.153\textwidth}
%         \centering
%     \includegraphics[width=\linewidth, height=4cm]{figures/mengguwen.png} 
%         \caption{Mongolian}
%         \label{fig:sub_b}
%     \end{subfigure}
%     \hfill
%     \begin{subfigure}{0.153\textwidth}
%         \centering
%         \includegraphics[width=\linewidth, height=4cm]{figures/zangwen.jpg} 
%         \caption{Tibetan}
%         \label{fig:sub_c}
%     \end{subfigure}
%      \vspace{-4mm}
%     \caption{Representative examples of low-resource multilingual historical documents, which contain important cultural, religious, and historical content. 
%     }
%      \vspace{-9mm}
%     \label{fig:three_side_by_side}
% \end{figure}

\section{Introduction}

% Historical document analysis is an important task for cultural heritage preservation, historical knowledge digitization, and low-resource language technology. 
% In recent years, optical character recognition (OCR) systems~\cite{wei2025deepseek,li2025dotsocrmultilingualdocumentlayout,dong2026mineru,wei2026deepseek,meng2019glyce} have achieved strong performance on modern printed documents and relatively clean historical document pages. 
% However, real historical documents remain substantially more challenging because the target text is often affected by severe physical degradation, including stroke fading, ink erosion, cracks, stains, wormholes, folds, and local structural breakage~\cite{ding2025hisdoc,zhang2025reviving,guan2025prep}. 
% These degradations do not merely reduce image quality; they directly destroy the visual evidence required for reliable character understanding, making historical document recognition fundamentally different from standard OCR~\cite{hegghammer2022ocr}.

Historical document analysis is important for cultural heritage preservation, historical knowledge digitization, and low-resource language technology. 
Recent optical character recognition (OCR) systems~\cite{wei2025deepseek,li2025dotsocrmultilingualdocumentlayout,dong2026mineru,wei2026deepseek,meng2019glyce} have achieved strong performance on modern printed documents and relatively clean historical pages. 
However, authentic historical documents remain substantially more challenging because the target text is frequently affected by severe physical degradation, including stroke fading, ink erosion, cracks, stains, wormholes, folds, and structural breakage~\cite{ding2025hisdoc,zhang2025reviving,guan2025prep}. 
Such degradation does not merely reduce visual quality, but can directly destroy the local evidence required for reliable character recognition, making historical document recognition fundamentally different from standard OCR~\cite{hegghammer2022ocr}.

A key challenge is that historical document pages usually contain mixed-visibility characters, ranging from visually clear to severely damaged or even illegible~\cite{pardo2024advanced}. 
This heterogeneous setting makes uniform visual recognition unreliable and tightly couples degradation perception with character recognition. 
A robust system must therefore estimate whether local visual evidence remains trustworthy before making character-level decisions. 
When the evidence is heavily corrupted, local recognition alone becomes insufficient and contextual reasoning is needed to recover missing content. 
However, existing methods typically emphasize either visual recognition or post-hoc restoration~\cite{assael2025contextualizing,assael2022restoring,prasetiadi2024fragmented}, and many pipelines apply contextual correction after recognition without explicitly considering character visibility~\cite{guan2025prep,kanerva2025ocr}. 
This strategy may unnecessarily overwrite correct predictions for visually clear characters. 
Ideally, clear characters should be recognized directly from local visual evidence, while contextual restoration should be selectively activated only for ambiguous or illegible regions.

This challenge is further amplified in low-resource multilingual historical documents. 
Existing public datasets and benchmarks remain largely concentrated on high-resource languages and widely studied scripts~\cite{zhang2025reviving,zhu2024reproducing,yang2025predicting}, whereas Tibetan, Mongolian, and Uyghur historical documents remain severely underexplored~\cite{aioanei2024deep}. 
Beyond limited annotations, these scripts exhibit distinct structural and directional characteristics: Tibetan contains stacked components, traditional Mongolian is written vertically with context-dependent glyph forms, and Uyghur follows right-to-left connected writing. 
Physical degradation can therefore disrupt stack composition, vertical continuity, and cursive connectivity, respectively, making recognition especially difficult under missing visual evidence. 
The scarcity of suitable benchmark resources further limits both fair evaluation and the development of models that jointly address degradation perception, character recognition, and contextual restoration in realistic low-resource settings.

% Furthermore, despite growing interest in historical document understanding, existing public datasets and benchmarks are still largely concentrated on high-resource languages, especially English and a few widely studied scripts~\cite{zhang2025reviving,zhu2024reproducing,yang2025predicting}. 
% In contrast, low-resource multilingual historical documents remain severely underexplored~\cite{aioanei2024deep}, particularly for scripts such as Tibetan, Mongolian, and Uyghur, where document degradation, limited annotations, and script-specific structural complexity coexist. 
% Compared with the predominantly horizontal recognition settings commonly studied for modern Chinese and English OCR, these scripts introduce additional structural and directional dependencies: Tibetan contains stacked components, traditional Mongolian is written vertically with context-dependent glyph forms, and Uyghur follows right-to-left connected writing. 
% Physical degradation can therefore disrupt stack composition, vertical continuity, or cursive connectivity, respectively, further increasing recognition difficulty and motivating our visibility-guided selective restoration.
% This lack of benchmark resources not only restricts fair evaluation, but also hinders the development of models capable of jointly handling degradation perception, character recognition, and contextual restoration in realistic low-resource settings.

To address these challenges, this paper presents \textbf{DDVR-Net}, a \textbf{D}egradation-aware \textbf{D}ecoupling and \textbf{V}isibility-guided \textbf{R}estoration framework for historical document recognition. 
Rather than treating the task as conventional OCR with auxiliary labels, DDVR-Net unifies degradation-aware perception, character recognition, and gated contextual restoration within a single framework. 
At a high level, the model first learns decoupled representations of glyph structure and degradation patterns, allowing it to separate intrinsic character identity from extrinsic physical corruption. 
It then performs local character recognition together with visibility prediction, so that each character region is assigned a reliability-aware state rather than being processed uniformly. 
For partially visible or severely degraded characters, DDVR-Net activates a visibility-guided contextual decoder to recover missing content from neighboring textual evidence, while preserving direct visual recognition for clear characters. 
In this way, contextual decoding serves as a selective fallback mechanism rather than a universal correction module, improving robustness on mixed-visibility historical documents. 
To support this task, we further construct a low-resource multilingual benchmark for degraded historical document recognition and restoration, covering Tibetan, Mongolian, and Uyghur historical documents. 
The benchmark provides character-level annotations, including bounding boxes, target labels, visibility states, and degradation types, enabling fine-grained evaluation of perception, recognition, and restoration in a unified setting. 
By focusing on low-resource multilingual scripts in historical documents rather than high-resource historical OCR scenarios, the benchmark helps address a current gap in public resources for this setting.
To facilitate future research and reproducibility, we will release the benchmark, inference code, and trained models to the community. 
In summary, the contributions of this work are three-fold:

(1) We construct a large-scale low-resource multilingual benchmark for historical document recognition and restoration, covering Tibetan, Mongolian, and Uyghur with 187000 document pairs and character-level visibility and degradation annotations.

(2) We formulate degraded historical document recognition as a reliability-conditioned perception-recognition-restoration problem and propose DDVR-Net, where predicted visibility controls contextual restoration. The contextual decoder selectively restores visually unreliable characters while preserving direct recognition for clear ones. 

(3) Extensive experiments and ablations validate the effectiveness of DDVR-Net, particularly its feature decoupling and visibility-guided restoration. The synthetic data further support effective synthetic-to-real transfer, achieving state-of-the-art benchmark performance and strong results on real-world subsets.

% (1) We construct a large-scale low-resource multilingual benchmark for historical document recognition and restoration, covering Tibetan, Mongolian, and Uyghur with 187,000 document pairs and character-level visibility and degradation annotations. We will release all generated and redistributable data, together with permissible annotations, metadata, and scripts for restricted sources.

% (2) We formulate degraded historical document recognition as a reliability-conditioned perception--recognition--restoration problem and propose DDVR-Net, where predicted visibility controls contextual restoration. The contextual decoder selectively restores visually unreliable characters while preserving recognition for clear ones. We will release the trained models and inference code.

% (3) Extensive experiments and ablations validate the effectiveness of DDVR-Net, particularly feature decoupling and visibility-guided restoration. The synthetic data also support effective synthetic-to-real transfer, yielding state-of-the-art benchmark performance and strong results on real-world subsets.

\section{Related Work}

\textbf{Historical Document Understanding and Restoration Methods}
Historical document analysis spans OCR~\cite{ham2025glyph}, layout parsing~\cite{zheng2026multimodal}, document restoration~\cite{zhao2025uni}, and context-based text recovery~\cite{sadeghian2024cvocr}. Recent methods improve historical document processing through semantic layout modeling~\cite{ding2025hisdoc}, visual restoration~\cite{yang2025predicting,zhang2025reviving}, and contextual reasoning~\cite{assael2025contextualizing,yu2025docthinker}. However, most approaches focus on either visual restoration, structural analysis, or textual completion, without explicitly estimating character-level visual reliability before contextual correction. Our work instead unifies perception, recognition, and restoration, using degradation-aware visibility to determine when local visual evidence should be preserved and when contextual restoration should be activated.

\noindent\textbf{Historical Document Datasets and Synthetic Benchmark Construction}
Historical document research has benefited from public benchmarks and synthetic data generation~\cite{nikolaidou2022survey,mazumder2023dataperf,bi2025llava}, including synthetic resources for epigraphy and historical OCR~\cite{aioanei2024deep,nacar2025sard} and benchmarks for damaged-document restoration~\cite{yang2025predicting,zhang2025reviving}. Nevertheless, existing resources remain concentrated on high-resource or relatively narrow script settings~\cite{shi2025large}. Low-resource historical documents, particularly Tibetan, Mongolian, and Uyghur, remain underrepresented with character-level visibility and degradation annotations. We therefore construct a multilingual benchmark covering these three scripts with bounding boxes, character labels, visibility states, and degradation types.

\section{Multilingual Historical Document Restoration Benchmark}

\begin{figure}[t]
    \centering
    \makebox[\linewidth][c]{%
        \includegraphics[width=1\linewidth]{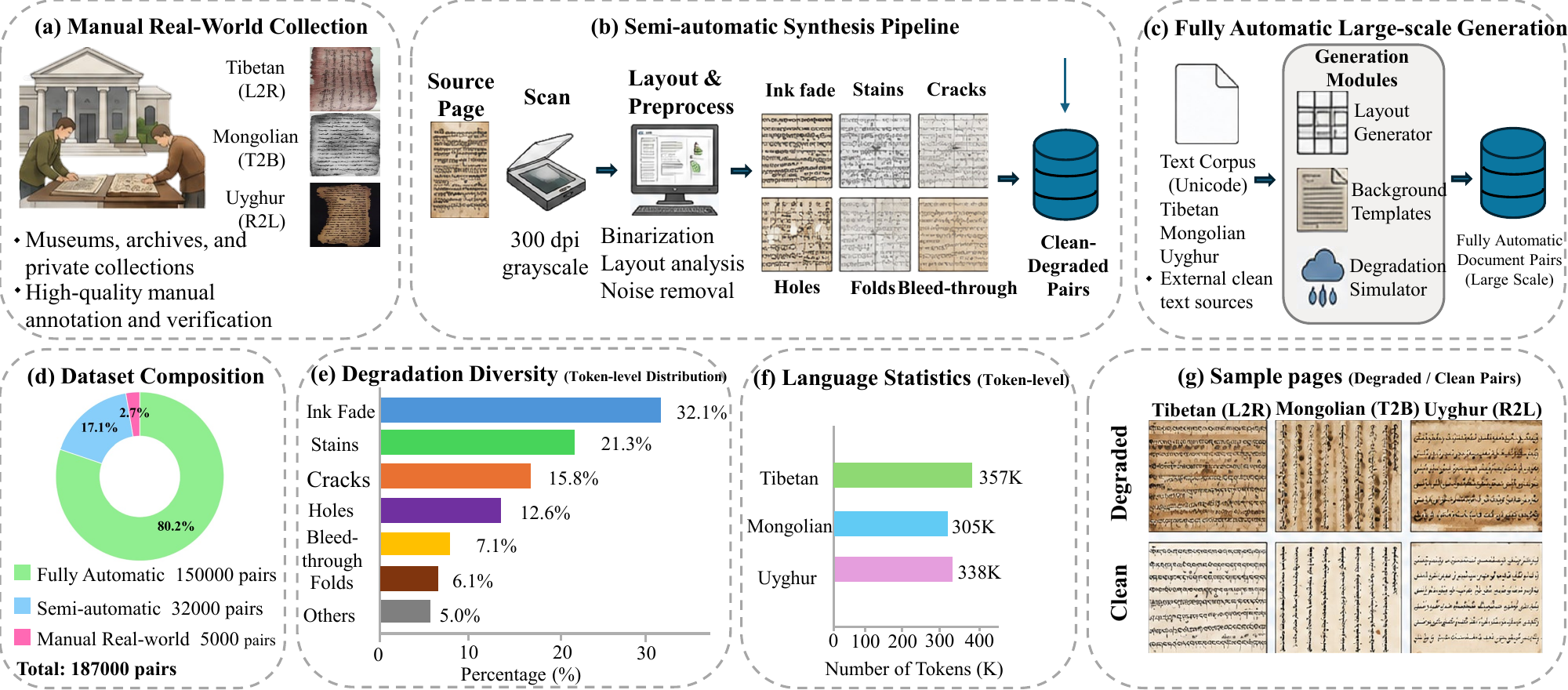}
    }
    \caption{Overview of benchmark construction.}
    \label{fig:pipeline}
    \vspace{-3mm}
\end{figure}

We construct a multilingual benchmark through three progressive pipelines, illustrated in Figure ~\ref{fig:pipeline}: (1) fully manual collection in real historical settings, (2) semi-automatic synthesis assisted by OCR tools, and (3) a fully automatic large-scale generation strategy. This design preserves the authenticity and diversity of real historical documents for evaluation, while providing sufficient scale and degradation variety for robust model training. The three scripts are intentionally selected to cover heterogeneous writing directions and glyph structures while sharing physical degradation mechanisms, rather than being treated as morphologically homogeneous.

\subsection{Professional Real-World Data Collection}
To ensure authenticity and historical validity, we collaborated with museums, archival institutions, and domain scholars who preserve representative documents of Tibetan, Mongolian, and Uyghur cultures~\footnote{Our annotation and validation protocol was informed by authoritative historical and archival resources, including CBETA, the National Library of China, the Rushi Tripitaka Collection, Tibetan manuscripts from Dunhuang and the Western Regions preserved in the British Library and the Bibliothèque nationale de France, the British Library collections, the Belgian Scheut Mission archives, and the National Library of China for Ethnic Minorities.}. These documents cover a wide range of genres, including Buddhist scriptures, historical chronicles, contracts, administrative records, and legal texts, spanning diverse aspects of history, politics, economy, society, and culture. This broad thematic coverage substantially enriches the diversity
of our benchmark and makes it a useful resource for studying
multilingual historical documents in realistic settings. Through this process, we collected 5000 real full-page document-text pairs. All real samples were annotated by native-speaking experts and doctoral researchers with prior experience in historical document analysis and restoration.

The annotation protocol adopts a character-level restoration approach. Annotators assign each character a content label, a visibility state
$v_i\in\{1,0,-1\}$ corresponding to \emph{clear},
\emph{partially degraded}, and \emph{severely degraded}, respectively,
and one or more degradation types $d_i$. Quality control involved independent double-annotation and expert resolution of disagreements. This real subset serves as high-quality supervision and a realistic evaluation benchmark.

\subsection{Large-Scale High-Fidelity Data Synthesis}
While real annotated documents provide high fidelity, their collection and verification remain labor-intensive and difficult to scale. We therefore expand the dataset with two complementary synthesis strategies, namely a semi-automatic pipeline built upon real historical pages and a fully automatic pipeline for large-scale generation.
\subsubsection{Semi-Automatic Synthesis Assisted by Historical Pages}

To overcome the scarcity of real clean-damaged page pairs, we construct synthetic data by applying controlled deterioration to well-preserved historical documents. This pipeline follows a reverse-restoration principle: progressively injecting historically plausible degradation into intact pages while preserving original layouts. Clean source pages from authenticated corpora are filtered (complete layout, $<3^\circ$ skew, high contrast) and normalized to 300 dpi grayscale. After localizing characters via transcription-aligned dots.ocr, we inject multi-level degradation. For character missing, LAMA removes strokes under 1--5 random irregular masks (covering 15\%--60\% of target boxes) per page. For paper damage, we overlay Bézier-generated cracks and holes, limiting corrupted areas to 2\%--12\% to prevent unrealistic over-destruction. For ink erosion and fading, genalog simulates blur, bleed-through, and stain diffusion using randomized kernels (3--11 pixels) and opacity attenuation (0.35--0.85). Finally, we render the text onto high-resolution historical paper textures and apply mild geometric perturbations (warping, fold shadows, and illumination drift). The resulting degraded-clean pairs are filtered using our quality control procedure (Section~\ref{sec:quality_assurance}), yielding 32000 high-quality semi-automatic synthetic pages.

\subsubsection{Fully Automatic Large-Scale Generation}

To scale the training data, we construct a fully automatic pipeline
for Tibetan, Mongolian, and Uyghur historical documents. Clean textual
corpora and representative page materials are collected from ethnic
museums, archival institutions, and university collaborators. From
real pages, we extract non-textual priors, including paper color,
fiber texture, border wear, stains, folds, line spacing, character
density, and ink intensity. We use
\texttt{gpt-5.2-pro-2025-12-11} through the API
(temperature $=0.0$) only to convert these priors into background
texture and style descriptions; it does not generate document text
or annotations. These descriptions condition SDXL-base-1.0 with a
LoRA adapter fine-tuned on authenticated historical paper textures
to synthesize blank backgrounds, which are normalized to 300-dpi
grayscale. Clean text is then rendered using script-specific historical
fonts and layout rules, with small perturbations in size, baseline,
spacing, and ink density. Exact prompts and generation parameters
are provided in Appendix~A.1.%\ref{app:synthesis_details}.

After obtaining clean page-text pairs, we inject degradation at both character and page levels. Character-level corruption includes partial stroke loss, blur, fading, and occlusion, implemented with irregular masks and pixel-level operations such as erosion, dilation, and contrast attenuation. Page-level corruption simulates cracks, wormholes, stains, folds, and paper damage through contour-based masks, SDXL-generated \cite{podell2024sdxl} textures, and alpha blending. Similar to the semi-automatic pipeline, these generated samples undergo the same quality assurance protocol. Using this pipeline, we synthesize approximately 50000 training samples for each language, resulting in 150000 fully automatic multilingual historical degradation-aware character recognition and restoration full-page pairs.

\subsubsection{Quality Assurance} \label{sec:quality_assurance}

To improve sample realism and remove low-quality cases, both the
semi-automatic and fully automatic pipelines undergo the same strict
two-stage quality-control procedure. First, rule-based screening
removes pages with excessive corruption or broken layouts, including
pages whose damaged area exceeds 25\%, whose foreground pixels fall
below 30\% of the clean source, or whose reading order is disrupted by
overlapping masks.

Second, each synthetic page is compared with real pages from the same language and document style. The CLIP and VLM thresholds were selected on a held-out validation subset by jointly considering visual realism and sample retention. We retain pages satisfying both a CLIP-based visual similarity above 0.82 and a VLM realism score above 4.0/5. The VLM evaluates paper texture, layout plausibility, degradation severity, and the absence of visually implausible artifacts. These scores are used as complementary filtering signals rather than as substitutes for human evaluation. To validate the retained samples, we conducted a blinded study in which real and synthetic backgrounds were randomly mixed and assessed by experts with over 25 years of experience in historical document preservation and analysis. Overall, 95\% of the retained synthetic backgrounds were judged historically plausible.

\subsection{Annotation and Data Distribution}

To support unified perception and restoration, each document image is annotated as $\mathcal{Y}=\{(b_i,c_i,v_i,d_i)\}_{i=1}^{N}$, where $b_i$ denotes the character bounding box, $c_i$ the target label, $v_i \in \{1,0,-1\}$ the visibility state (clear, partially degraded, and severely degraded, respectively), and $d_i$ the degradation category. To reduce evaluation leakage, we perform data splitting at the document-page level, such that all characters from the same page belong to only one split. The manual subset is document-disjoint, and the semi-automatic subset is source-disjoint, i.e., all degraded variants derived from the same source page are assigned to a single split. In addition, pages in the manual test split are excluded from the semi-automatic source pool, and synthetic text is screened to reduce exact or near-duplicate overlap with the real evaluation set. All manual samples are annotated by native-speaking annotators and independently verified by two annotators, with disagreements resolved through expert adjudication; agreement statistics and the adjudication protocol are provided in Appendix~A.3. Following this protocol, the manual, semi-automatic, and fully automatic subsets are partitioned into training and test sets with a 7:3 ratio, as summarized in Table~\ref{tab:multilingual_stats}.

\begin{table}[t]
\centering
\caption{Statistics of the multilingual historical document restoration dataset. ``Dam./Im'' denotes average damaged characters per image, ``Tot. Dam.'' the total damaged characters, and ``Cls.'' the number of character categories.}
\vspace{-3mm}
\label{tab:multilingual_stats}
\renewcommand{\arraystretch}{1} % 稍微调紧行高
\setlength{\tabcolsep}{1.5pt} % 极限压缩列与列之间的空白
\footnotesize % 将表格基础字号设为 footnotesize
\resizebox{\columnwidth}{!}{% 此时表格本身变窄了，resizebox 缩小的比例会大幅减小，字体就大了
\begin{tabular}{@{}llccccccc@{}}
\toprule
\textbf{Lang.} & \textbf{Subset} & \textbf{Imgs} & \textbf{Dam./Im} & \textbf{Tot. Dam.} & \textbf{Char/Im} & \textbf{Cls.} &\textbf{Train} &\textbf{Test} \\ \midrule

\multirow{3}{*}{Tibetan}
& Manual & 2000 & 105.40 & 210800 & 512.35 & 1120 & 1400 & 600 \\
& Semi-Auto & 12000 & 62.15 & 745800 & 485.20 & 3250 & 8400 & 3600 \\
& Fully Auto & 50000 & 58.20 & 2910000 & 498.60 & 3480 & 35000 & 15000 \\ \midrule

\multirow{3}{*}{Mongolian}
& Manual & 1500 & 88.65 & 132975 & 355.40 & 650 & 1050 & 450 \\
& Semi-Auto & 10000 & 45.30 & 453000 & 320.15 & 1840 & 7000 & 3000 \\
& Fully Auto & 50000 & 42.80 & 2140000 & 332.40 & 1950 & 35000 & 15000 \\ \midrule

\multirow{3}{*}{Uyghur}
& Manual & 1500 & 94.20 & 141300 & 438.90 & 720 & 1050 & 450 \\
& Semi-Auto & 10000 & 52.80 & 528000 & 410.60 & 1560 & 7000 & 3000 \\
& Fully Auto & 50000 & 50.60 & 2530000 & 426.80 & 1680 & 35000 & 15000 \\ \midrule

\rowcolor[HTML]{EFEFEF}
\textbf{Total} & \textbf{Combined} & \textbf{187000} & \textbf{52.36} & \textbf{9791875} & \textbf{418.37} & \textbf{7440} & \textbf{130900} & \textbf{56100} \\ \bottomrule
\end{tabular}}
\vspace{-3mm}
\end{table}

% Furthermore, the three subsets naturally support a progressive curriculum learning strategy from synthetic to real data. Specifically, the model is first pretrained on the large-scale \textit{fully automatic} subset to learn general character representations and degradation-aware visual patterns. It is then further adapted on the \textit{semi-automatic} subset, which provides historically more plausible layouts and degradation characteristics. Finally, the model is fine-tuned on the \textit{manual} training split to improve performance under real historical document conditions. This progressive curriculum effectively alleviates the synthetic-to-real domain gap and improves the model's generalization ability in practical restoration scenarios.

\section{Methodology}
\begin{figure*}[t]
    \centering
    \includegraphics[width=\textwidth]{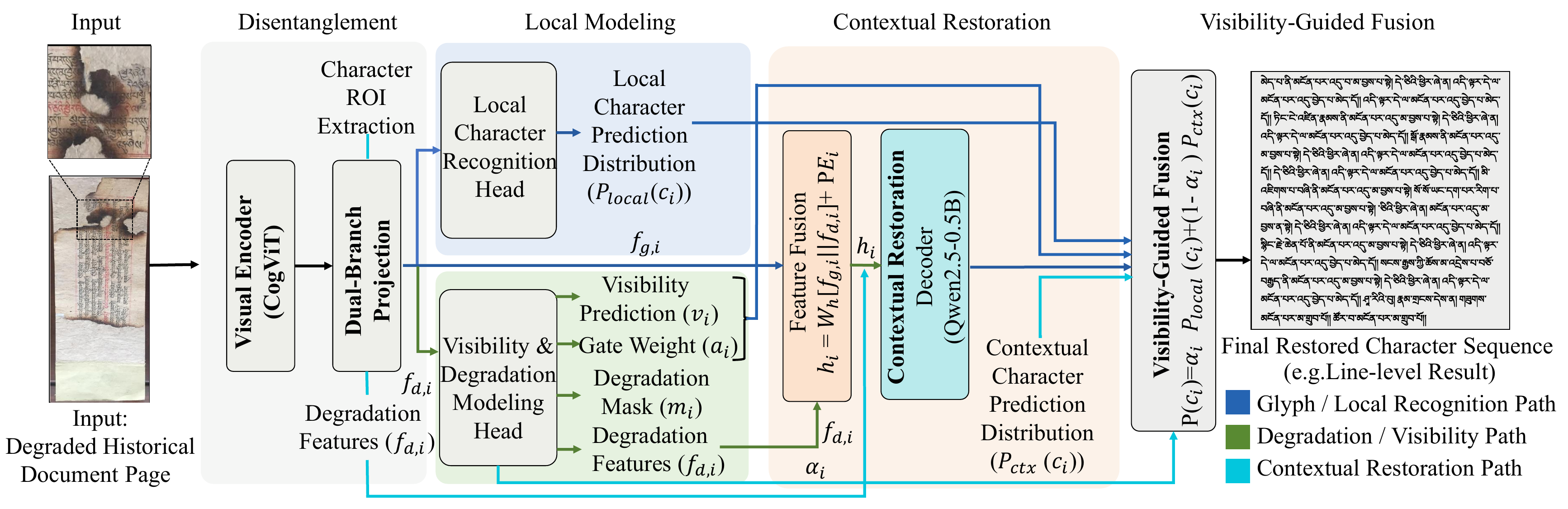}
    %\vspace{-7mm}
    \caption{Overall framework of DDVR-Net.}
    \label{fig:framework}
    % \vspace{-2mm}
\end{figure*}

Figure~\ref{fig:framework} illustrates the pipeline of DDVR-Net. The methodological contribution does not rely on treating the
individual building blocks as novel in isolation. Instead, DDVR-Net
couples degradation representation, visibility estimation, local
recognition, and contextual decoding through a
reliability-conditioned decision mechanism. Visibility determines
whether local evidence should be preserved or contextual restoration
should be activated, thereby addressing the trade-off between missing
visual evidence and language-model over-correction. The framework follows a three-stage design: (1) degradation-aware representation decoupling,
(2) local perception and reliability modeling, and
(3) degradation-conditioned contextual restoration.
Given a degraded historical document page and its character-region
proposals, the model first disentangles glyph and degradation features,
then predicts local character identities and visibility states, and
finally performs selective contextual restoration for unreliable
characters through visibility-guided fusion. Our experiments
use annotated character boxes to isolate recognition and restoration
from upstream detection; this controlled protocol does not assume that
perfect boxes are always available in deployment. In an end-to-end setting, the annotated boxes can be replaced by
proposals from an external text detector, and the effect of proposal
noise is analyzed in Appendix~C.5.5.

\subsection{Degradation-Aware Representation Learning}
\label{StageI}
Instead of relying on a standard multi-task framework that directly maps degraded image patches to character labels, the first part of our framework introduces a \textbf{Degradation-Aware Representation Learning} paradigm. The primary objective is to explicitly disentangle the intrinsic semantic information (the character glyph) from extrinsic environmental noise (the degradation patterns) at the feature level. 
Given an input document image $I$ and a sequence of $N$ character
regions $\mathcal{B}=(b_1,b_2,\ldots,b_N)$, together with character
labels $\mathcal{C}=(c_1,c_2,\ldots,c_N)$, visibility states
$v_i\in\{-1,0,1\}$, and multi-hot degradation labels $d_i$, we first
extract high-level global spatial contexts. During training and the
controlled recognition evaluation, $\mathcal{B}$ denotes annotated
character boxes. For end-to-end inference, it may instead denote
detector-generated proposals $\widehat{\mathcal{B}}$; inaccurate or
missing proposals affect the RoI features and are therefore evaluated
separately through controlled perturbations. We employ CogViT (0.4B)~\cite{duan2026glm} as our Visual Encoder to process the entire image, producing a comprehensive global feature map $F = Enc_{vis}(I)$. Subsequently, we utilize the RoIAlign operation to pool and extract fixed-size local feature representations $f_i$ for each specific character region $b_i$. To achieve feature disentanglement, we hypothesize that a character's inherent topological structure remains constant regardless of its physical degradation, treating the deterioration simply as an external visual perturbation. We design a \textbf{Decoupled Representation Learning} module comprising two parallel branches: a Glyph Branch and a Degradation Branch. Both branches are instantiated as Multi-Layer Perceptrons (MLPs) that project the flattened region-of-interest feature $f_i$ into two latent spaces. Formally, this dual-branch mapping is defined as $f_{g,i} = G_g(f_i)$ and $f_{d,i} = G_d(f_i)$, where $f_{g,i}$ is intended to capture stroke topologies, structural components, and stylistic geometries, while $f_{d,i}$ is intended to capture the morphological characteristics of physical damages like cracks, stains, or occlusions. To enforce this disentanglement, we formulate a composite decoupling objective $\mathcal{L}_{decouple}$ that concurrently optimizes glyph consistency, degradation discriminability, and feature orthogonality.

\textbf{(1) Glyph Consistency:} We apply Supervised Contrastive Learning (SupCon) to the glyph representations $f_{g,i}$. The goal is to pull the representations of the same character class closer together in the latent space even if they suffer from entirely different degradation types while pushing apart representations of different characters. Crucially, severely degraded or entirely invisible characters (where $v_i = -1$) provide unreliable and noisy structural signals. Therefore, we introduce a visibility mask to filter out these instances during this stage, relying on \ref{StageIII} for their contextual restoration. The masked glyph consistency loss is defined in
Equation~(1):
{
\begin{equation}
\label{eq:glyph_loss} % 可选：添加标签方便后续在正文中用 \ref{eq:glyph_loss} 引用
\mathcal{L}_{glyph} = \sum_{i=1}^{N}(v_i \neq -1) \left[ - \frac{1}{|P(i)|} \sum_{p \in P(i)} \log \frac{\exp(sim(f_{g,i}, f_{g,p}) / \tau)}{\sum_{j \neq i} \exp(sim(f_{g,i}, f_{g,j}) / \tau)} \right]
\end{equation}
}where $P(i)$ denotes the set of indices for positive samples sharing the same character label $c_i$ within a batch, $sim(\cdot, \cdot)$ is the cosine similarity, and $\tau$ is the temperature scaling parameter. \textbf{(2) Degradation Discriminability:} Because a single historical character frequently suffers from multiple deteriorations (e.g., moth-eaten holes and water stains), we model the degradation perception as a multi-label classification task. We append a linear classification head $W_{deg}$ to the degradation branch and optimize it using a Binary Cross-Entropy (BCE) loss against the multi-hot degradation target $d_i$: $\mathcal{L}_{deg}=BCEWithLogits(W_{deg}\cdot f_{d,i},d_i)$. This ensures that $f_{d,i}$ encodes highly discriminative patterns regarding the specific types of physical damage in the local region. \textbf{(3) Feature Orthogonality:} Finally, independently optimizing the two branches does not guarantee mutual exclusivity; the glyph branch might still encode subtle degradation noise, and vice versa. To enforce a strict structural-noise decoupling, we introduce a feature orthogonality penalty. By minimizing squared cosine similarity between the L2-normalized glyph and degradation vectors, we encourage latent representations to capture complementary visual properties:
{
\begin{equation}
    \mathcal{L}_{ortho} = \frac{1}{N} \sum_{i=1}^{N} \left\| \frac{f_{g,i}^T f_{d,i}}{\|f_{g,i}\|_2 \|f_{d,i}\|_2} \right\|_F^2
\end{equation}
}
\vspace{-3mm}
\subsection{Multi-task Local Perception and Degradation Modeling}
\label{StageII}
Building on the decoupled latent spaces, we model local semantics and character reliability through two parallel tasks: Local Character Recognition and Visibility Prediction.

\textbf{(1) Local Character Recognition:} Since the glyph representation $f_{g,i}$ has been disentangled from environmental noise, we utilize it exclusively for local character recognition. We pass $f_{g,i}$ through a recognition head $Head_{rec}$ to obtain the local visual prediction distribution $P_{local}(c_i) \in \mathbb{R}^{|C|}$, where $|C|$ is the vocabulary size. The objective is optimized using the standard Cross-Entropy loss: $\mathcal{L}_{rec} = \sum_{i=1}^{N}(v_i \neq -1) CE(P_{local}(c_i), c_i)$. Note that we mask out severely degraded characters ($v_i = -1$) during this local supervision, as their visual features lack sufficient structural integrity to reliably predict the character class without contextual aid. \textbf{(2) Visibility Prediction:} Concurrently, the degradation representation $f_{d,i}$ inherently captures the physical condition of the character. We leverage $f_{d,i}$ to predict the visibility state $\hat{v}_i \in \{-1, 0, 1\}$ via a visibility head $Head_{vis}$. This task models the reliability of the local visual cue and is
optimized by Equation~(3):
{
\begin{equation}
    \mathcal{L}_{vis} = \sum_{i=1}^{N} CE(Head_{vis}(f_{d,i}), v_i)
\end{equation}
}
\subsection{Degradation-Conditioned Contextual Restoration}
\label{StageIII}
We argue that forcing a language model to predict visually unambiguous characters is redundant and often leads to over-correction. Instead, contextual reasoning should act as a fallback mechanism, strictly triggered only when local visual evidence is compromised. To achieve this, we formulate a given text line as a sequence of character representations $H = (h_1, h_2, \dots, h_N)$. Each token $h_i$ is constructed by fusing the decoupled local visual features, explicit degradation features, and spatial positional encodings (PE): $h_i = W_h [f_{g,i} \parallel f_{d,i}] + PE_i$, where $W_h$ is a linear projection layer aligning the visual dimension to the language model's latent space. We employ a lightweight LLM as our restoration decoder $Enc_{ctx}$. The sequence $H$ serves as visual prompts fed into the decoder to capture bidirectional textual semantics and adjacent character contexts: $\tilde{H} = Enc_{ctx}(H)$ where $\tilde{H} = (\tilde{h}_1, \tilde{h}_2, \dots, \tilde{h}_N)$ represents the contextually enriched sequence. The contextual restoration distribution is then computed via a linear projection and Softmax activation: $P_{ctx}(c_i) = Softmax(W_{ctx} \tilde{h}_i)$ Crucially, to prevent the decoder from “blind guessing" and to enforce degradation-conditioned learning, we introduce a deterministic degradation mask $m_i$. Contextual restoration is exclusively activated for partially degraded ($v_i = 0$) or severely degraded/invisible ($v_i = -1$) characters:
{
\begin{equation}
    m_i = \begin{cases} 1, & \text{if } v_i \in \{0, -1\} \\ 0, & \text{otherwise} \end{cases}
\end{equation}
}The contextual restoration loss in Equation (5) is constrained
by the mask defined in Equation (4).
{
\begin{equation}
    \mathcal{L}_{restore} = \sum_{i=1}^{N} m_i \cdot CE(P_{ctx}(c_i), c_i)
\end{equation}
} Finally, the unified recognition probability dynamically merges the local and contextual distributions via the visibility-guided gate $\alpha_i$ (computed in \ref{StageII}). For clear characters ($m_i=0$), $\alpha_i \to 1$, relying entirely on local vision; for degraded characters ($m_i=1$), $\alpha_i \to 0$, seamlessly shifting the reliance to $P_{ctx}(c_i)$: $P(c_i) = \alpha_i P_{local}(c_i) + (1 - \alpha_i) P_{ctx}(c_i)$, where $\alpha_i \in [0,1]$ denotes the visibility-guided gating weight predicted from the reliability modeling branch in \ref{StageII}. A larger $\alpha_i$ indicates that the model places higher confidence on local visual recognition, whereas a smaller $\alpha_i$ encourages stronger reliance on contextual restoration. Here, both $P_{local}(c_i)$ and $P_{ctx}(c_i)$ are normalized probability distributions over the character vocabulary. Based on the fused prediction, the final recognition loss is defined as $
\mathcal{L}_{final} = \sum_{i=1}^{N} CE(P(c_i), c_i).
$ This design ensures that prediction is supervised after visibility-guided fusion, improving consistency between local perception and contextual restoration.

\vspace{-3mm}
\subsection{Three-Stage Training Strategy}

Directly optimizing our heterogeneous framework (visual encoder coupled with a language decoder) from scratch often leads to instability. To address this, we adopt a coarse-to-fine progressive training strategy: \textit{representation decoupling}, \textit{local perception learning}, and \textit{contextual restoration alignment}. This design ensures that the contextual decoder is introduced
after the visual branch has learned priors; configurations are provided in
Appendix~B.1.

\noindent \textbf{Stage 1: Decoupled Visual Representation Learning.} We first optimize the visual encoder and dual-branch projections to disentangle character identity from degradation. The glyph branch uses contrastive consistency, the degradation branch uses classification, and both are regularized by orthogonality. The objective is $\mathcal{L}_{Stage\_I} = \lambda_1 \mathcal{L}_{glyph} + \lambda_2 \mathcal{L}_{deg} + \lambda_3 \mathcal{L}_{ortho}$, establishing noise-free character representations and explicit degradation contexts.

\noindent \textbf{Stage 2: Local Perception and Reliability Modeling.} Next, we train the local recognition and visibility prediction heads on the decoupled features. Following Section 4.2, severely degraded characters are excluded from local recognition supervision, while all participate in visibility prediction. The objective becomes $\mathcal{L}_{Stage\_II} = \mathcal{L}_{Stage\_I} + \lambda_4 \mathcal{L}_{rec} + \lambda_5 \mathcal{L}_{vis}$, equipping the model with local semantics and a reliability indicator.

\noindent \textbf{Stage 3: Contextual Restoration Alignment.} Finally, retaining the visual backbone as the primary perception component, we unfreeze only the projection layer and the language decoder, fine-tuning the latter with LoRA~\cite{hu2022lora}. Contextual restoration is supervised exclusively on partially or severely degraded characters to prevent over-correcting visually clear text. The final training objective is $\mathcal{L}_{Stage\_III} = \mathcal{L}_{Stage\_II} + \lambda_6 \mathcal{L}_{restore} + \lambda_7 \mathcal{L}_{final}$.

\begin{figure*}[t] % 如果是双栏论文想跨两栏显示，用 figure*；单栏用 figure
  \centering
  \begin{subfigure}{0.33\textwidth}
    \centering
    \includegraphics[width=4.5cm, height=3cm]{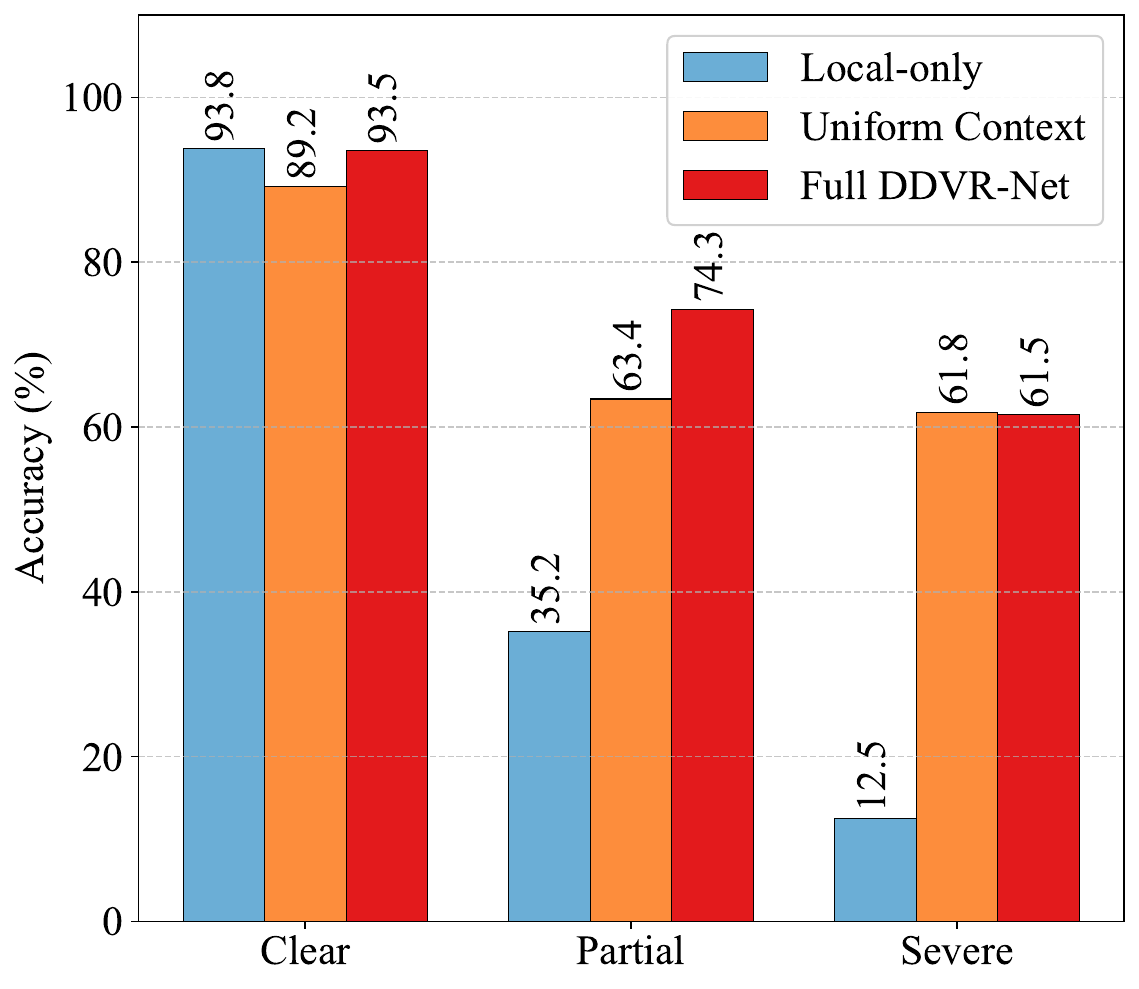}
    \caption{Performance Stratified by Visibility}
    \label{fig:sub1}
  \end{subfigure}
  \hfill % 在子图之间插入弹性间距
  \begin{subfigure}{0.33\textwidth}
    \centering
    \includegraphics[width=4.5cm, height=3cm]{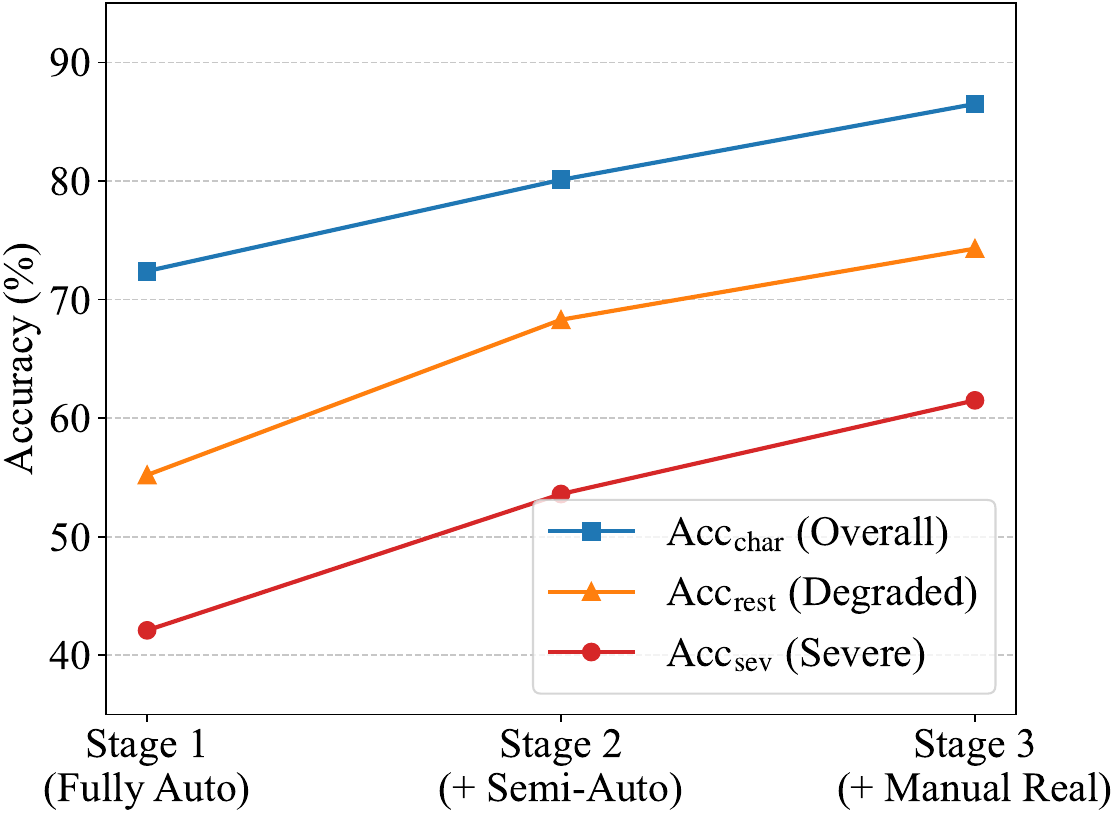}
    \caption{Effect of Progressive Curriculum}
    \label{fig:sub2}
  \end{subfigure}
  \hfill
  \begin{subfigure}{0.33\textwidth}
    \centering
    \includegraphics[width=4.5cm, height=3cm]{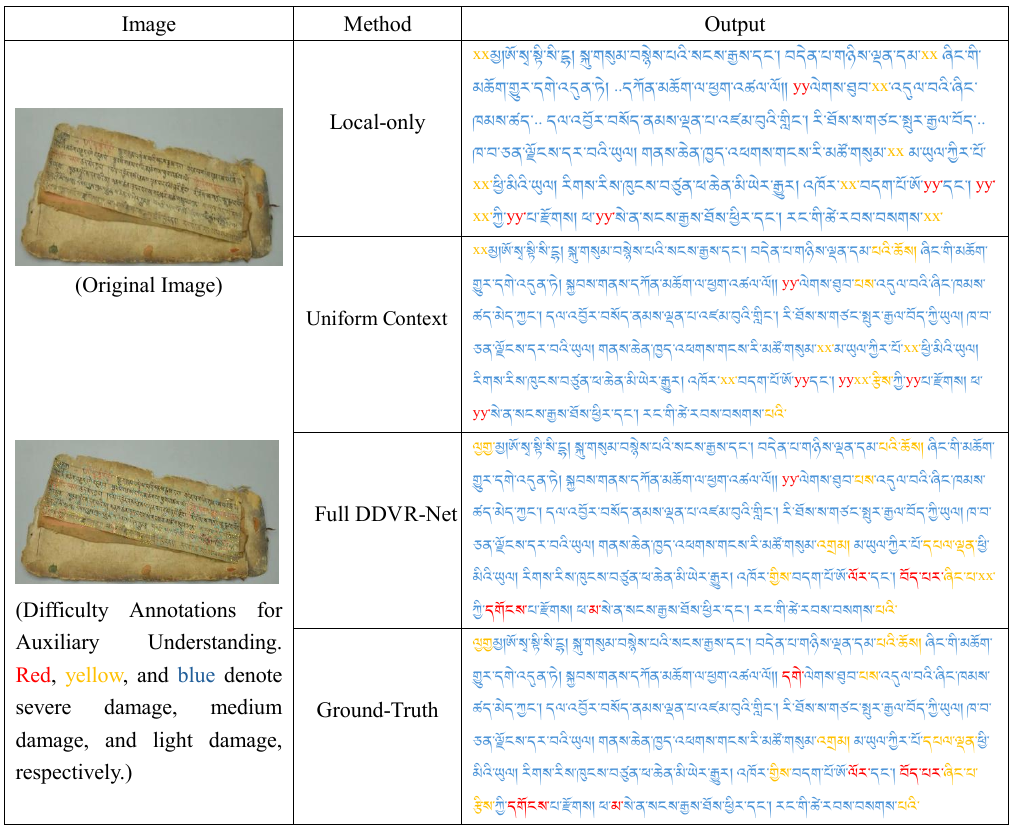} 
    \caption{Mixed-visibility text comparison.}
    \label{fig:sub3}
  \end{subfigure}
  \vspace{-3mm}
 \caption{Analysis of DDVR-Net: (a) visibility stratification, (b) progressive curriculum, and (c) case study.}
  \label{fig:total_figure}
\end{figure*}

\section{Experiments}

\subsection{Implementation Detail}

We implement DDVR-Net in PyTorch, utilizing CogViT (0.4B) as the visual encoder and Qwen2.5-0.5B-Instruct (fine-tuned via LoRA~\cite{hu2022lora}) as the restoration decoder. Input images are normalized to 300 dpi grayscale. We train on 32 NVIDIA A100 GPUs for 100 epochs (total batch size 32) using AdamW~\cite{loshchilov2017decoupled} ($\beta_{1,2}=(0.9, 0.999)$) with a base learning rate of $1 \times 10^{-4}$ and cosine annealing. To bridge the synthetic-to-real gap, our curriculum progressively pre-trains on the 150K fully automatic pairs, adapts on 32K semi-automatic pairs, and fine-tunes on the manual real-world split. Finally, the multi-task loss balancing coefficients
$\lambda_{1:7}$ are empirically set to
$\{1.0, 0.5, 0.1, 1.0, 0.5, 0.8, 1.0\}$; additional implementation
details are provided in Appendix~C.1.1.

\subsection{Evaluation Metrics and Baselines}
We use the character-level metrics defined in
Appendix~C.2, while respecting native
reading orders (Tibetan L2R, Uyghur R2L, and Mongolian T2B). \textbf{Recognition \& Restoration:} $\mathrm{Acc}_{\mathrm{char}}$, $\mathrm{Acc}_{\mathrm{rest}}$, and $\mathrm{Acc}_{\mathrm{sev}}$ measure accuracy on all, degraded ($v_i \in \{\texttt{partial}, \texttt{severe}\}$), and severe ($v_i = \texttt{severe}$) characters, respectively. $\mathrm{CER}$ is the Levenshtein based error rate over sequences $\mathcal{S}$. \textbf{Sequence-Level:} $\mathrm{LEM}$ and $\mathrm{SeqAcc}$ evaluate exact match for lines $\mathcal{L}$ and pages $\mathcal{P}$. \textbf{Visibility \& Degradation:} Category-wise visibility accuracy $\mathrm{Acc}_{c}$ is computed for $c \in \{\texttt{clear},$ $\texttt{partial},$ $\texttt{severe}\}$. Degradation perception is a multi-label task evaluated via $\mathrm{Micro\mbox{-}F1}$ and $\mathrm{Macro\mbox{-}F1}$ across $K$ categories using $\mathrm{TP}_k, \mathrm{FP}_k, \mathrm{FN}_k$.

To validate our dataset and DDVR-Net architecture, we compare against strong baselines across three categories: \textbf{(1) MLLMs}: We evaluate generalist models including GPT-4o~\cite{achiam2023gpt},
Gemini 3.1 Pro~\cite{team2024gemini}, and Qwen3-VL~\cite{bai2025qwen3} uniformly via
their official APIs; the unified prompts, decoding settings, and
output-parsing protocol are provided in
Appendix~C.3. \textbf{(2) Document Parsing and OCR Systems}: We benchmark against state-of-the-art OCR models such as dots.ocr~\cite{li2025dotsocrmultilingualdocumentlayout}, PaddleOCR 3.0~\cite{cui2025paddleocr}, and GLM-OCR~\cite{duan2026glm}. \textbf{(3) Historical Document Recognition Methods}: We compare against recent works for degraded texts: Reviving Cultural Heritage~\cite{zhang2025reviving} and PreP-OCR~\cite{guan2025prep}. Since these were primarily evaluated on Chinese documents, we re-train them from scratch on our multilingual benchmark splits using their original hyperparameter configurations. This aligns the training data and original configurations across task-specific baselines; the controlled same-backbone comparisons in Section~5.4.1 further isolate the effect of DDVR-Net's selective restoration design.

\subsection{Main Results}

\begin{table*}[t]
\centering
\caption{Overall results of different models on \textbf{Tibetan, Mongolian, and Uyghur} historical documents.}
\vspace{-3mm}
\label{tab:multilingual_results}
\setlength{\tabcolsep}{5pt} % 稍微收缩列间距以适应宽度
\renewcommand{\arraystretch}{1} % 稍微调紧行高
\resizebox{\linewidth}{!}{
\begin{tabular}{c|c|cccc|cc|ccc|cc}
\toprule
\multirow{2}{*}{\textbf{Lang.}} & \multirow{2}{*}{\textbf{Models}} 
& \multicolumn{4}{c|}{\textbf{Character Recognition \& Restoration}}  
& \multicolumn{2}{c|}{\textbf{Sequence-Level}} 
& \multicolumn{3}{c|}{\textbf{Visibility Estimation}} 
& \multicolumn{2}{c}{\textbf{Degradation-Type Classification}} \\
\cmidrule(lr){3-6}\cmidrule(lr){7-8} \cmidrule(lr){9-11} \cmidrule(lr){12-13}
& & $\mathrm{Acc}_{\mathrm{char}}$ 
& $\mathrm{Acc}_{\mathrm{rest}}$ 
& $\mathrm{Acc}_{\mathrm{sev}}$ 
& $\mathrm{CER} \downarrow$ 
& $\mathrm{LEM}$ 
& $\mathrm{SeqAcc}$ 
& $\mathrm{Acc}_{\mathrm{clear}}$ 
& $\mathrm{Acc}_{\mathrm{partial}}$ 
& $\mathrm{Acc}_{\mathrm{severe}}$ 
& $\mathrm{Micro\mbox{-}F1}$ 
& $\mathrm{Macro\mbox{-}F1}$ \\
\midrule

% --- Tibetan ---
\multirow{9}{*}{\rotatebox{90}{\textbf{Tibetan}}} 
& GPT-4o              & 58.72 & 38.41 & 22.15 & 28.43 & 12.36 &  7.84 & 66.51 & 47.62 & 32.75 & 46.85 & 44.29 \\
& Gemini 3.1 Pro      & 57.84 & 37.66 & 21.43 & 29.15 & 11.89 &  7.41 & 65.26 & 46.19 & 31.84 & 45.38 & 43.12 \\
& Qwen3-VL            & 61.29 & 42.18 & 25.84 & 26.87 & 14.72 &  9.65 & 68.14 & 49.37 & 34.22 & 48.73 & 46.51 \\
\cmidrule{2-13}
& dots.ocr             & 52.34 & 28.15 & 12.39 & 45.67 &  4.51 &  2.18 & 60.12 & 37.53 & 21.65 & 36.42 & 33.87 \\
& PaddleOCR 3.0    & 55.42 & 32.47 & 15.63 & 42.14 &  6.82 &  3.59 & 63.85 & 41.26 & 24.81 & 40.19 & 37.43 \\
& GLM-OCR             & 53.76 & 30.68 & 14.21 & 43.89 &  5.73 &  2.94 & 62.37 & 39.84 & 23.46 & 38.71 & 35.62 \\
\cmidrule{2-13}
& Reviving Cultural Heritage & 68.45 & 51.27 & 36.82 & 18.76 & 24.53 & 16.89 & 72.48 & 58.12 & 47.93 & 61.35 & 58.84 \\
& PreP-OCR                 & 72.18 & 55.34 & 40.15 & 16.42 & 28.71 & 19.42 & 75.63 & 62.81 & 51.46 & 64.72 & 62.39 \\
\cmidrule{2-13}
& \textbf{DDVR-Net (Ours)} & \textbf{86.47} & \textbf{74.29} & \textbf{61.54} & \textbf{9.83} & \textbf{45.62} & \textbf{36.75} & \textbf{89.31} & \textbf{78.46} & \textbf{69.18} & \textbf{81.24} & \textbf{79.67} \\
\midrule

% --- Mongolian ---
\multirow{9}{*}{\rotatebox{90}{\textbf{Mongolian}}} 
& GPT-4o              & 55.62 & 35.81 & 20.34 & 31.57 &  9.83 &  6.19 & 63.48 & 44.95 & 29.71 & 43.52 & 41.86 \\
& Gemini 3.1 Pro      & 54.38 & 34.69 & 19.52 & 32.41 &  9.14 &  5.76 & 62.15 & 43.82 & 28.63 & 42.17 & 40.54 \\
& Qwen3-VL            & 58.14 & 39.52 & 23.17 & 29.83 & 11.45 &  7.28 & 64.92 & 46.71 & 31.84 & 45.61 & 43.25 \\
\cmidrule{2-13}
& dots.ocr             & 49.63 & 25.18 & 10.45 & 49.36 &  2.71 &  1.12 & 57.39 & 34.27 & 18.64 & 33.58 & 31.25 \\
& PaddleOCR 3.0    & 52.81 & 29.56 & 13.24 & 45.19 &  4.15 &  2.38 & 61.27 & 38.64 & 22.51 & 37.15 & 34.93 \\
& GLM-OCR             & 50.15 & 27.42 & 11.83 & 47.62 &  3.48 &  1.57 & 59.74 & 36.51 & 20.19 & 35.82 & 33.46 \\
\cmidrule{2-13}
& Reviving Cultural Heritage & 65.73 & 48.15 & 33.62 & 21.49 & 21.87 & 14.36 & 69.51 & 55.28 & 44.93 & 58.46 & 56.12 \\
& PreP-OCR                 & 69.84 & 52.61 & 37.56 & 18.95 & 25.13 & 16.58 & 73.42 & 60.19 & 48.74 & 61.85 & 59.47 \\
\cmidrule{2-13}
& \textbf{DDVR-Net (Ours)} & \textbf{84.35} & \textbf{71.18} & \textbf{58.42} & \textbf{11.23} & \textbf{41.76} & \textbf{32.49} & \textbf{87.61} & \textbf{75.84} & \textbf{66.27} & \textbf{78.53} & \textbf{76.92} \\
\midrule

% --- Uyghur ---
\multirow{9}{*}{\rotatebox{90}{\textbf{Uyghur}}} 
& GPT-4o              & 56.49 & 36.72 & 21.14 & 30.18 & 10.57 &  6.81 & 64.73 & 45.86 & 30.62 & 44.91 & 42.75 \\
& Gemini 3.1 Pro      & 55.73 & 35.84 & 20.36 & 31.45 &  9.82 &  6.37 & 63.58 & 44.71 & 29.54 & 43.64 & 41.38 \\
& Qwen3-VL            & 59.26 & 40.85 & 24.63 & 28.51 & 12.64 &  8.42 & 66.41 & 48.27 & 33.15 & 47.29 & 45.16 \\
\cmidrule{2-13}
& dots.ocr             & 50.84 & 26.79 & 11.23 & 47.58 &  3.46 &  1.72 & 58.62 & 35.94 & 20.37 & 34.85 & 32.53 \\
& PaddleOCR 3.0    & 54.16 & 31.54 & 14.85 & 43.72 &  5.91 &  3.14 & 62.49 & 40.13 & 23.78 & 38.74 & 36.19 \\
& GLM-OCR             & 52.37 & 28.91 & 13.56 & 45.24 &  4.83 &  2.45 & 60.83 & 38.45 & 21.92 & 37.16 & 34.68 \\
\cmidrule{2-13}
& Reviving Cultural Heritage & 66.89 & 49.36 & 34.72 & 20.83 & 22.45 & 15.62 & 70.84 & 56.73 & 46.18 & 59.72 & 57.45 \\
& PreP-OCR                 & 70.43 & 53.82 & 38.47 & 18.65 & 26.84 & 17.51 & 74.25 & 61.38 & 50.29 & 63.47 & 60.82 \\
\cmidrule{2-13}
& \textbf{DDVR-Net (Ours)} & \textbf{85.34} & \textbf{72.87} & \textbf{59.61} & \textbf{10.45} & \textbf{43.58} & \textbf{34.29} & \textbf{88.54} & \textbf{76.92} & \textbf{67.35} & \textbf{79.46} & \textbf{77.83} \\
\bottomrule
\end{tabular}
}
\vspace{-3mm}
\end{table*}

Table~\ref{tab:multilingual_results} shows that DDVR-Net consistently achieves state-of-the-art performance across Tibetan, Mongolian, and Uyghur documents. Unlike conventional OCR or general MLLMs, DDVR-Net excels in the most challenging metrics, validating our core innovations. By decoupling intrinsic glyph structures from extrinsic noise, the model accurately perceives local reliability. The visibility-guided gate prevents over-correcting legible text by selectively triggering the contextual decoder only for ambiguous regions, supporting stable restoration across different reading orders. Furthermore, stable gains across the three languages suggest that the benchmark provides a useful testbed for cross-language evaluation. The performance on real-world test sets indicates that our progressive pipeline (manual curation combined with realism-oriented synthesis) successfully bridges the synthetic-to-real gap. High scores in auxiliary tasks (visibility and degradation classification) further suggest that fine-grained character-level supervision is crucial. These results demonstrate that low-resource historical document understanding benefits from a unified perception-recognition-restoration framework beyond visual recognition alone. We further evaluate DDVR-Net on the fully automatic,
semi-automatic, and manual real-world test subsets. Subset-specific results are provided in
Appendix~C.4.

\subsection{Ablation Study}
\subsubsection{Fair Baselines under the Same Backbone and Data}

\begin{table}[t]
\centering
\caption{Fair same-backbone comparison across languages.}
\label{tab:fair_baselines_lang}
\renewcommand{\arraystretch}{1} % 稍微调紧行高
\setlength{\tabcolsep}{4pt}
\resizebox{\columnwidth}{!}{
\begin{tabular}{c l ccccc}
\toprule
\multirow{2}{*}{\textbf{Lang.}} 
& \multirow{2}{*}{\textbf{Training Paradigm}} 
& \multicolumn{4}{c}{\textbf{Character Recognition \& Restoration}} 
& \textbf{Sequence-Level} \\
\cmidrule(lr){3-6} \cmidrule(lr){7-7}
& 
& $\mathrm{Acc}_{\mathrm{char}}$ 
& $\mathrm{Acc}_{\mathrm{rest}}$ 
& $\mathrm{Acc}_{\mathrm{sev}}$ 
& $\mathrm{CER} \downarrow$ 
& $\mathrm{SeqAcc}$ \\
\midrule

\multirow{5}{*}{\rotatebox[origin=c]{90}{\textbf{Tibetan}}}
& \textbf{Full DDVR-Net} & {\LARGE \textbf{86.47}} & {\LARGE \textbf{74.29}} & {\LARGE \textbf{61.54}} & {\LARGE \textbf{9.83}} & {\LARGE \textbf{36.75}} \\
& Confidence-based Correction & {\LARGE 85.10} & {\LARGE 73.60} & {\LARGE 62.10} & {\LARGE 10.90} & {\LARGE 32.60} \\
& Post-hoc Correction & {\LARGE 84.80} & {\LARGE 72.90} & {\LARGE 59.80} & {\LARGE 11.20} & {\LARGE 31.80} \\
& Uniform Context & {\LARGE 84.20} & {\LARGE 72.80} & {\LARGE 60.90} & {\LARGE 11.70} & {\LARGE 29.40} \\
& Local-only & {\LARGE 83.90} & {\LARGE 68.90} & {\LARGE 54.60} & {\LARGE 12.80} & {\LARGE 30.70} \\
\midrule

\multirow{5}{*}{\rotatebox[origin=c]{90}{\textbf{Mongolian}}}
& \textbf{Full DDVR-Net} & {\LARGE \textbf{84.35}} & {\LARGE \textbf{71.18}} & {\LARGE \textbf{58.42}} & {\LARGE \textbf{11.23}} & {\LARGE \textbf{32.49}} \\
& Confidence-based Correction & {\LARGE 82.40} & {\LARGE 70.40} & {\LARGE 58.60} & {\LARGE 12.20} & {\LARGE 28.70} \\
& Post-hoc Correction & {\LARGE 82.00} & {\LARGE 69.90} & {\LARGE 56.90} & {\LARGE 12.60} & {\LARGE 27.90} \\
& Uniform Context & {\LARGE 81.20} & {\LARGE 69.70} & {\LARGE 57.30} & {\LARGE 13.00} & {\LARGE 25.80} \\
& Local-only & {\LARGE 80.70} & {\LARGE 65.40} & {\LARGE 50.80} & {\LARGE 14.10} & {\LARGE 26.90} \\
\midrule

\multirow{5}{*}{\rotatebox[origin=c]{90}{\textbf{Uyghur}}}
& \textbf{Full DDVR-Net} & {\LARGE \textbf{85.34}} & {\LARGE \textbf{72.87}} & {\LARGE \textbf{59.61}} & {\LARGE \textbf{10.45}} & {\LARGE \textbf{34.29}} \\
& Confidence-based Correction & {\LARGE 83.70} & {\LARGE 72.20} & {\LARGE 60.30} & {\LARGE 11.50} & {\LARGE 30.40} \\
& Post-hoc Correction & {\LARGE 83.20} & {\LARGE 71.60} & {\LARGE 58.40} & {\LARGE 11.80} & {\LARGE 29.60} \\
& Uniform Context & {\LARGE 82.50} & {\LARGE 71.10} & {\LARGE 58.90} & {\LARGE 12.30} & {\LARGE 27.50} \\
& Local-only & {\LARGE 81.90} & {\LARGE 66.80} & {\LARGE 52.70} & {\LARGE 13.40} & {\LARGE 28.60} \\
\bottomrule
\end{tabular}}
\vspace{-3mm}
\end{table}

To verify that the gains of DDVR-Net come from its selective restoration design rather than stronger backbone capacity or additional data, we compare it with controlled baselines under the same visual encoder, contextual decoder, training data, and schedule. As shown in Table~\ref{tab:fair_baselines_lang}, although \textbf{(1) Confidence-based Correction}, \textbf{(2) Post-hoc Correction}, and \textbf{(3) Uniform Context} all improve over \textbf{(4) Local-only}, they remain consistently weaker than DDVR-Net across Tibetan, Mongolian, and Uyghur. In particular, Uniform Context improves $\mathrm{Acc}_{rest}$ and $\mathrm{Acc}_{sev}$ but reduces $\mathrm{SeqAcc}$, revealing over-correction, while Confidence-based Correction shows that local confidence alone is an unreliable proxy for degradation visibility. Under identical visual encoders, contextual decoders, training data,
and schedules, these results isolate the contribution of the proposed
reliability-conditioned mechanism: the gain arises from using
visibility to determine when and how strongly contextual restoration
should be activated, rather than from merely combining existing
modules or adding a language decoder. Additional baseline definitions and language-wise
analyses are provided in Appendix~C.5.1.

\subsubsection{Effect of Representation Decoupling}
\label{sec:ablation_decoupling}

To verify that gains stem from disentanglement, we ablate Stage I (Table~\ref{tab:ablation_decoupling}): (1) \textbf{w/o Decoupling}: removes the dual-branch design; (2) \textbf{w/o Ortho}: removes $\mathcal{L}_{ortho}$; (3) \textbf{w/o Glyph}: removes $\mathcal{L}_{glyph}$; (4) \textbf{w/o Deg}: removes $\mathcal{L}_{deg}$ (detailed settings and analyses are provided in
Appendix~C.5.2). Results show that \textbf{w/o Decoupling} exhibits the sharpest decline in $\mathrm{Acc}_{\mathrm{sev}}$ and CER, as the model misidentifies deterioration as glyph structures. Each constraint contributes to performance: \textbf{w/o Ortho}
increases subspace leakage between the two latent spaces, while
\textbf{w/o Glyph Consistency} and \textbf{w/o Deg Cls} weaken
recognition consistency and visibility estimation, respectively. Thus, the ablation results suggest that Stage I decoupling contributes to reliable degradation-conditioned restoration.

\begin{table}[t]
\centering
\caption{Ablation: Representation Decoupling}
\label{tab:ablation_decoupling}
\renewcommand{\arraystretch}{1} % 稍微调紧行高
\setlength{\tabcolsep}{4pt}
\resizebox{\linewidth}{!}{
\begin{tabular}{c|l|cccc|cc|cc}
\toprule
\multirow{2}{*}{\textbf{Lang.}} 
& \multirow{2}{*}{\textbf{Models}} 
& \multicolumn{4}{c|}{\textbf{Character Recognition \& Restoration}}  
& \multicolumn{2}{c|}{\textbf{Visibility Estimation}} 
& \multicolumn{2}{c}{\textbf{Degradation-Type Classification}} \\
\cmidrule(lr){3-6}\cmidrule(lr){7-8}\cmidrule(lr){9-10}
& 
& \textbf{$\mathrm{Acc}_{\mathrm{char}}$} 
& \textbf{$\mathrm{Acc}_{\mathrm{rest}}$} 
& \textbf{$\mathrm{Acc}_{\mathrm{sev}}$} 
& \textbf{$\mathrm{CER} \downarrow$} 
& \textbf{$\mathrm{Acc}_{\mathrm{partial}}$} 
& \textbf{$\mathrm{Acc}_{\mathrm{severe}}$} 
& \textbf{$\mathrm{Micro\mbox{-}F1}$} 
& \textbf{$\mathrm{Macro\mbox{-}F1}$} \\
\midrule

% Tibetan
\multirow{5}{*}{\rotatebox[origin=c]{90}{\textbf{Tibetan}}} 
& \textbf{Full DDVR-Net} & {\LARGE \textbf{86.47}} & {\LARGE \textbf{74.29}} & {\LARGE \textbf{61.54}} & {\LARGE \textbf{9.83}}  & {\LARGE \textbf{78.46}} & {\LARGE \textbf{69.18}} & {\LARGE \textbf{81.24}} & {\LARGE \textbf{79.67}} \\
& w/o Decoupling        & {\LARGE 78.50} & {\LARGE 58.20} & {\LARGE 42.30} & {\LARGE 15.50} & {\LARGE 62.10} & {\LARGE 51.40} & {\LARGE 65.20} & {\LARGE 63.10} \\
& w/o Ortho             & {\LARGE 84.10} & {\LARGE 70.50} & {\LARGE 57.20} & {\LARGE 10.90} & {\LARGE 75.30} & {\LARGE 66.10} & {\LARGE 77.80} & {\LARGE 76.20} \\
& w/o Glyph Consistency & {\LARGE 82.30} & {\LARGE 68.40} & {\LARGE 55.10} & {\LARGE 11.80} & {\LARGE 76.50} & {\LARGE 67.40} & {\LARGE 79.10} & {\LARGE 78.00} \\
& w/o Deg Cls           & {\LARGE 83.50} & {\LARGE 64.20} & {\LARGE 48.60} & {\LARGE 13.20} & {\LARGE 58.40} & {\LARGE 46.20} & - & - \\ 
\midrule

% Mongolian
\multirow{5}{*}{\rotatebox[origin=c]{90}{\textbf{Mongolian}}} 
& \textbf{Full DDVR-Net} & {\LARGE \textbf{84.35}} & {\LARGE \textbf{71.18}} & {\LARGE \textbf{58.42}} & {\LARGE \textbf{11.23}} & {\LARGE \textbf{75.84}} & {\LARGE \textbf{66.27}} & {\LARGE \textbf{78.53}} & {\LARGE \textbf{76.92}} \\
& w/o Decoupling        & {\LARGE 76.21} & {\LARGE 54.36} & {\LARGE 39.15} & {\LARGE 17.48} & {\LARGE 59.42} & {\LARGE 48.11} & {\LARGE 62.15} & {\LARGE 60.34} \\
& w/o Ortho             & {\LARGE 82.14} & {\LARGE 67.85} & {\LARGE 54.62} & {\LARGE 12.35} & {\LARGE 72.16} & {\LARGE 63.45} & {\LARGE 75.28} & {\LARGE 73.81} \\
& w/o Glyph Consistency & {\LARGE 80.52} & {\LARGE 65.41} & {\LARGE 52.18} & {\LARGE 13.56} & {\LARGE 73.85} & {\LARGE 64.12} & {\LARGE 76.42} & {\LARGE 74.95} \\
& w/o Deg Cls           & {\LARGE 81.36} & {\LARGE 61.25} & {\LARGE 46.58} & {\LARGE 14.82} & {\LARGE 55.34} & {\LARGE 44.27} & - & - \\ 
\midrule

% Uyghur
\multirow{5}{*}{\rotatebox[origin=c]{90}{\textbf{Uyghur}}} 
& \textbf{Full DDVR-Net} & {\LARGE \textbf{85.34}} & {\LARGE \textbf{72.87}} & {\LARGE \textbf{59.61}} & {\LARGE \textbf{10.45}} & {\LARGE \textbf{76.92}} & {\LARGE \textbf{67.35}} & {\LARGE \textbf{79.46}} & {\LARGE \textbf{77.83}} \\
& w/o Decoupling        & {\LARGE 77.45} & {\LARGE 56.12} & {\LARGE 40.85} & {\LARGE 16.21} & {\LARGE 60.58} & {\LARGE 49.62} & {\LARGE 63.74} & {\LARGE 61.92} \\
& w/o Ortho             & {\LARGE 83.15} & {\LARGE 69.45} & {\LARGE 55.82} & {\LARGE 11.64} & {\LARGE 73.54} & {\LARGE 64.51} & {\LARGE 76.12} & {\LARGE 74.85} \\
& w/o Glyph Consistency & {\LARGE 81.28} & {\LARGE 67.14} & {\LARGE 53.46} & {\LARGE 12.78} & {\LARGE 75.12} & {\LARGE 65.84} & {\LARGE 77.65} & {\LARGE 75.91} \\
& w/o Deg Cls           & {\LARGE 82.41} & {\LARGE 63.15} & {\LARGE 47.38} & {\LARGE 14.05} & {\LARGE 57.26} & {\LARGE 45.18} & - & - \\
\bottomrule
\end{tabular}
}
\vspace{-3mm}
\end{table}

\subsubsection{Effect of Visibility-Guided Selective Restoration}
\label{sec:ablation_visibility}

To justify selective restoration, we evaluate DDVR-Net against four variants in Table~\ref{tab:ablation_visibility}: (1) \textbf{Uniform Context}: applies the decoder to all characters; (2) \textbf{w/o Visibility Head}: removes prediction; (3) \textbf{Hard Switch}: uses a binary threshold; (4) \textbf{Local-only}: removes the decoder (detailed settings and analyses are provided in
Appendix~C.5.3). Results in Figure~\ref{fig:sub1} reveal a clear trade-off: Local-only fails on degraded regions ($\mathrm{Acc}_{\mathrm{sev}}$) due to lacking context, while \textbf{Uniform Context} over-corrects clear text, dropping $\mathrm{Acc}_{\mathrm{clear}}$ and SeqAcc. Hard Switch consistently underperforms the continuous gate, indicating that posterior-conditioned soft fusion better accommodates ambiguous and boundary-visibility characters than a discrete inference rule. Thus, DDVR-Net acts as a targeted fallback, learning when to activate the contextual decoder and how strongly to rely on it.

\subsubsection{Synthetic-to-Real Generalization and Data Composition}
\label{sec:ablation_curriculum}
To validate our stage-wise integration of 187K document pairs, we compare five training paradigms in Table~\ref{tab:ablation_curriculum}: (1) \textbf{Manual Only}: uses only 5K real pairs; (2) \textbf{w/o Manual}: omits real-world fine-tuning; (3) \textbf{Mixed}: combines all data simultaneously; (4) \textbf{Reverse}: reverses the stage order; and (5) \textbf{Full Curriculum}: our proposed auto-to-manual path (detailed settings and analyses are provided in
Appendix~C.5.4). Results in Figure~\ref{fig:sub2} show that
\textbf{Manual Only} performs substantially worse under the limited
real-data regime, while \textbf{w/o Manual} retains a clear
synthetic-to-real domain gap. Our Full Curriculum succeeds by using automatic data for large-scale priors, semi-automatic data for layout bridging, and manual data for calibration, ensuring robust performance under realistic conditions.

\subsubsection{Robustness to Imperfect Character Boxes.}
The main experiments use annotated character boxes to isolate the
recognition restoration problem. To quantify error propagation from
an upstream detector, we additionally perturb box coordinates and
randomly remove a fraction of character proposals. The robustness protocol for imperfect upstream character boxes,
including coordinate perturbations and missing proposals, is described
in Appendix~C.5.5.

\subsection{Case Study}
\label{sec:case_study}
Figure~\ref{fig:sub3} qualitatively compares
restoration strategies on a mixed-visibility historical document page;
additional examples and analyses are provided in
Appendix~C.6.
For clear characters (blue box), Local-only and DDVR-Net succeed, whereas Uniform Context over-corrects visual evidence via hallucinations. In severely degraded regions (red box), Local-only fails, while DDVR-Net dynamically triggered by estimated visibility accurately restores semantics, matching Uniform Context. These results suggest that DDVR-Net better handles mixed-visibility characters by selectively using contextual information without degrading clear-text recognition.

\begin{table}[t]
\centering
\caption{Effect of Visibility-Guided Selective Restoration}
\label{tab:ablation_visibility}
\setlength{\tabcolsep}{4pt}
\renewcommand{\arraystretch}{1}
\resizebox{\linewidth}{!}{
\footnotesize
\begin{tabular}{c|l|cccc|cc|c}
\toprule
\multirow{2}{*}{\textbf{Lang.}}
& \multirow{2}{*}{\textbf{Models}}
& \multicolumn{4}{c|}{\textbf{Character Recognition \& Restoration}}
& \multicolumn{2}{c|}{\textbf{Sequence-Level}}
& \multicolumn{1}{c}{\textbf{Clear Chars}} \\
\cmidrule(lr){3-6}\cmidrule(lr){7-8}\cmidrule(lr){9-9}
&
& $\mathrm{Acc}_{\mathrm{char}}$
& $\mathrm{Acc}_{\mathrm{rest}}$
& $\mathrm{Acc}_{\mathrm{sev}}$
& $\mathrm{CER} \downarrow$
& $\mathrm{LEM}$
& $\mathrm{SeqAcc}$
& $\mathrm{Acc}_{\mathrm{clear-char}}$ \\
\midrule

% Tibetan
\multirow{5}{*}{\rotatebox[origin=c]{90}{\scriptsize \textbf{Tibetan}}}
& \textbf{Full DDVR-Net}            & \textbf{86.47} & 74.29 & 61.54 & \textbf{9.83} & \textbf{45.62} & \textbf{36.75} & 93.50 \\
& Uniform Context for All          & 84.10 & \textbf{74.50} & \textbf{61.80} & 11.50 & 42.10 & 33.50 & 89.20 \\
& w/o Visibility Head              & 83.20 & 68.50 & 55.40 & 12.30 & 40.50 & 31.80 & 91.00 \\
& Hard Switch instead of Soft Gate & 85.10 & 71.80 & 58.90 & 10.60 & 43.80 & 34.90 & 92.80 \\
& Local-only                       & 75.40 & 35.20 & 12.50 & 22.40 & 25.60 & 18.50 & \textbf{93.80} \\
\midrule

% Mongolian
\multirow{5}{*}{\rotatebox[origin=c]{90}{\scriptsize \textbf{Mongolian}}}
& \textbf{Full DDVR-Net}            & \textbf{84.35} & 71.18 & 58.42 & \textbf{11.23} & \textbf{41.76} & \textbf{32.49} & 92.10 \\
& Uniform Context for All          & 82.50 & \textbf{71.50} & \textbf{58.60} & 12.80 & 38.50 & 29.80 & 87.50 \\
& w/o Visibility Head              & 81.40 & 66.30 & 53.20 & 13.90 & 37.20 & 28.50 & 90.10 \\
& Hard Switch instead of Soft Gate & 83.20 & 69.10 & 56.30 & 12.10 & 40.10 & 31.00 & 91.50 \\
& Local-only                       & 73.80 & 33.50 & 10.80 & 24.10 & 22.40 & 15.20 & \textbf{92.40} \\
\midrule

% Uyghur
\multirow{5}{*}{\rotatebox[origin=c]{90}{\scriptsize \textbf{Uyghur}}}
& \textbf{Full DDVR-Net}            & \textbf{85.34} & 72.87 & 59.61 & \textbf{10.45} & \textbf{43.58} & \textbf{34.29} & 93.10 \\
& Uniform Context for All          & 83.20 & \textbf{73.10} & \textbf{59.80} & 12.10 & 40.20 & 31.50 & 88.40 \\
& w/o Visibility Head              & 82.10 & 67.50 & 54.80 & 13.10 & 38.50 & 30.10 & 90.80 \\
& Hard Switch instead of Soft Gate & 84.10 & 70.40 & 57.50 & 11.20 & 41.80 & 32.80 & 92.20 \\
& Local-only                       & 74.50 & 34.10 & 11.50 & 23.50 & 24.10 & 16.80 & \textbf{93.30} \\
\bottomrule
\end{tabular}
}
\vspace{-3mm}
\end{table}

\begin{table}[t]
\centering
\caption{Effect of Progressive Curriculum over Heterogeneous Data Sources (Evaluated on Manual Real-World Test Set)}
\label{tab:ablation_curriculum}
\setlength{\tabcolsep}{4pt}
\renewcommand{\arraystretch}{1}
\resizebox{\linewidth}{!}{
\footnotesize
\begin{tabular}{c|l|ccccc}
\toprule
\multirow{2}{*}{\textbf{Lang.}}
& \multirow{2}{*}{\textbf{Training Paradigm}}
& \multicolumn{4}{c}{\textbf{Character Recognition \& Restoration}}
& \textbf{Sequence-Level} \\
\cmidrule(lr){3-6} \cmidrule(lr){7-7}
&
& $\mathrm{Acc}_{\mathrm{char}}$
& $\mathrm{Acc}_{\mathrm{rest}}$
& $\mathrm{Acc}_{\mathrm{sev}}$
& $\mathrm{CER} \downarrow$
& $\mathrm{SeqAcc}$ \\
\midrule

% Tibetan
\multirow{5}{*}{\rotatebox[origin=c]{90}{\scriptsize \textbf{Tibetan}}}
& \textbf{Full Curriculum (Auto $\rightarrow$ Semi $\rightarrow$ Manual)} & \textbf{86.47} & \textbf{74.29} & \textbf{61.54} & \textbf{9.83} & \textbf{36.75} \\
& Mixed Training (Random Shuffle)                                   & 83.15 & 69.80 & 56.20 & 12.10 & 32.40 \\
& Reverse Curriculum (Manual $\rightarrow$ Semi $\rightarrow$ Auto) & 80.40 & 65.30 & 52.10 & 14.50 & 28.50 \\
& w/o Manual Fine-tune (Auto $\rightarrow$ Semi)                    & 81.20 & 66.80 & 54.50 & 13.80 & 29.80 \\
& Manual Only                                                       & 76.50 & 58.40 & 41.20 & 18.20 & 22.10 \\
\midrule

% Mongolian
\multirow{5}{*}{\rotatebox[origin=c]{90}{\scriptsize \textbf{Mongolian}}}
& \textbf{Full Curriculum (Auto $\rightarrow$ Semi $\rightarrow$ Manual)} & \textbf{84.35} & \textbf{71.18} & \textbf{58.42} & \textbf{11.23} & \textbf{32.49} \\
& Mixed Training (Random Shuffle)                                   & 80.90 & 66.50 & 53.10 & 13.50 & 28.10 \\
& Reverse Curriculum (Manual $\rightarrow$ Semi $\rightarrow$ Auto) & 78.20 & 62.10 & 49.30 & 16.10 & 24.30 \\
& w/o Manual Fine-tune (Auto $\rightarrow$ Semi)                    & 79.10 & 63.80 & 51.20 & 15.40 & 25.80 \\
& Manual Only                                                       & 74.30 & 55.20 & 38.50 & 20.50 & 18.40 \\
\midrule

% Uyghur
\multirow{5}{*}{\rotatebox[origin=c]{90}{\scriptsize \textbf{Uyghur}}}
& \textbf{Full Curriculum (Auto $\rightarrow$ Semi $\rightarrow$ Manual)} & \textbf{85.34} & \textbf{72.87} & \textbf{59.61} & \textbf{10.45} & \textbf{34.29} \\
& Mixed Training (Random Shuffle)                                   & 82.10 & 68.20 & 54.80 & 12.80 & 30.10 \\
& Reverse Curriculum (Manual $\rightarrow$ Semi $\rightarrow$ Auto) & 79.50 & 64.10 & 50.40 & 15.20 & 26.50 \\
& w/o Manual Fine-tune (Auto $\rightarrow$ Semi)                    & 80.30 & 65.50 & 52.70 & 14.30 & 27.80 \\
& Manual Only                                                       & 75.80 & 57.10 & 39.80 & 19.10 & 20.20 \\
\bottomrule
\end{tabular}
}
\end{table}

\section{Conclusion}
We propose DDVR-Net, a unified perception-recognition-restoration framework for historical document analysis under severe degradation and mixed-visibility characters. DDVR-Net decouples intrinsic glyph structures from extrinsic corruption and employs visibility-guided selective restoration to recover damaged regions while avoiding over-correction of clear text. To address data scarcity, we introduce a high-fidelity synthesis pipeline and a 187000-pair multilingual benchmark across Tibetan, Mongolian, and Uyghur scripts. This resource helps bridge the synthetic-to-real gap while supporting reliable and secure digital preservation of low-resource historical documents. Extensive experiments show that DDVR-Net achieves the strongest results among the compared methods on the proposed benchmark, including document-disjoint and source-disjoint manual real-world test sets. Evaluation on independent third-party benchmarks remains future work due to the limited availability of comparable character-level resources for degraded historical Tibetan, Mongolian, and Uyghur documents.

% \section{Conclusion}
% We propose DDVR-Net, a unified perception-recognition-restoration framework for historical document analysis, addressing challenges posed by severe degradation and mixed-visibility characters. DDVR-Net decouples intrinsic glyph structures from extrinsic corruption, employing a visibility-guided selective restoration that recovers damaged regions while avoiding over-correction of clear text. To overcome data scarcity, we introduce a high-fidelity synthesis pipeline and a 187000-pair multilingual benchmark across Tibetan, Mongolian, and Uyghur scripts. This resource enables the model to learn robust degradation priors and bridge the synthetic-to-real gap without post-processing. Extensive experiments show that DDVR-Net achieves the strongest results
% among the compared methods on the proposed benchmark, including its
% document-disjoint and source-disjoint manual real-world test sets.
% Evaluation on independent third-party benchmarks remains future work
% because directly comparable character-level resources for degraded
% historical Tibetan, Mongolian, and Uyghur documents are currently
% limited.

\section*{Acknowledgments}

This work was supported by the Graduate Research Projects of Minzu University of China.

\bibliographystyle{ACM-Reference-Format}
\balance
\bibliography{sample-base}

% 开始附录
\appendix
[sigconf]acmart
{booktabs}
{multirow}
{array}
{makecell}
{tabularx}
{colortbl}
{xcolor}
{graphicx}
{subcaption}
{multirow}
{booktabs}
{makecell}
{graphicx}
{subcaption}
{listings}

\settopmatter{
  printacmref=false,
  printccs=false,
  printfolios=true
}

\renewcommand\footnotetextcopyrightpermission[1]{}

\makeatletter
\fancyhf{}
\pagestyle{plain}
\makeatother

% \title{DDVR-Net: Degradation-Aware Decoupling and Visibility-Guided Restoration for Historical Document Recognition}
% \maketitle
\section*{Appendices}

\makeatletter
\fancyhf{}
\thispagestyle{plain}
\makeatother

\section{Multilingual Historical Document Restoration Benchmark}
\label{app:benchmark}

\subsection{Fully Automatic Synthesis Details}
\label{app:synthesis_details}

\subsubsection{GPT-Assisted Texture Description}
\label{app:gpt_texture}
We used gpt-5.2-pro-2025-12-11 through the official API with a temperature of 0.0 solely to convert non-textual priors extracted from real historical document pages into concise texture and style descriptions. GPT was not used to generate document images, textual content, character labels, or annotations. For each language and collection, the following prompt template was used: \textit{``Describe a blank historical paper background for [LANGUAGE] documents from [COLLECTION] using only non-textual visual attributes. The description should specify the paper color, fiber texture, surface roughness, border wear, stains, folds, aging marks, and illumination conditions. Do not include any characters, words, symbols, seals, illustrations, or other readable content. Return only a concise texture and style description suitable for conditioning a historical paper background generator.''} The resulting descriptions were subsequently used as text conditions for SDXL-base-1.0 \cite{podell2024sdxl} with a LoRA adapter fine-tuned on authenticated historical paper textures to synthesize blank page backgrounds.

\subsubsection{SDXL and LoRA Configuration}
\label{app:sdxl_lora}
We used SDXL-base-1.0 to synthesize blank historical paper backgrounds
conditioned on the GPT-generated texture descriptions. SDXL inference
used the DPM++ 2M Karras sampler with 30 denoising steps, a
classifier-free guidance scale of 7.0, a resolution of
$1024\times1024$, and a fixed random seed of 42. The LoRA adapter was
configured with a rank of 16 and an alpha value of 16, and was trained
for 10,000 steps using a learning rate of $1\times10^{-4}$. The LoRA
training data contained only authenticated blank or text-masked
historical paper textures, preventing textual content or annotations
from being introduced during background synthesis.

\subsection{Annotation and Data Distribution}
\label{app:annotation_distribution}

To support unified degradation perception, character recognition, and
contextual restoration, each document page is annotated as
$\mathcal{Y}=\{(b_i,c_i,v_i,d_i)\}_{i=1}^{N}$, where $N$ denotes the
number of characters on the page. Specifically, $b_i$ is the character
bounding box, $c_i$ is the target character label,
$v_i\in\{1,0,-1\}$ represents the visibility state
(\emph{clear}, \emph{partially degraded}, and
\emph{severely degraded}, respectively), and $d_i$ denotes one or more
degradation categories. This character-level annotation supports
joint supervision for recognition, visibility estimation, degradation
classification, and context-aware restoration.

To prevent evaluation ambiguity and content leakage, all data are split
at the \textbf{document-page level}, rather than at the character or
line level. Consequently, all characters from the same page belong to
only one partition. For the manually collected subset, the split is
further constrained to be \textbf{document-disjoint}: pages originating
from the same historical source document cannot appear in both training
and test sets. For the semi-automatic subset, all degraded variants
derived from the same source page are assigned to a single partition,
making the training and test sets \textbf{source-disjoint}. For the
fully automatic subset, the partition is determined before final
rendering, and pages are then independently generated from sampled
text, layout priors, and background templates to reduce near-duplicate
cross-partition samples.

We additionally prevent leakage across the three construction
pipelines. The source pool used for semi-automatic synthesis is
maintained separately from the manually annotated real-world test set;
in particular, no page in the manual test split is reused as a source
for semi-automatic generation. For the fully automatic subset, text is
sampled from external clean corpora and template resources, followed
by exact- and near-duplicate screening against the real-world
evaluation split. These procedures minimize both
\textbf{document-level overlap} and \textbf{textual overlap} between
synthetic training data and authentic evaluation pages, reducing the
possibility that performance gains arise from memorized content.

Annotation quality is controlled through independent double annotation
and senior-expert adjudication. Each manually collected page is
annotated by two trained native-speaking annotators with experience in
historical document analysis. Disagreements in character transcription,
visibility state, character boundary, or degradation category are
resolved according to a fixed annotation guideline. Detailed
inter-annotator agreement statistics and the adjudication protocol are
reported in Appendix~\ref{app:annotation_agreement}.

Following these principles, the manual, semi-automatic, and fully
automatic subsets are partitioned into training and test sets with a
7:3 ratio. The manual training data expose the model to authentic
historical distributions, while the document-disjoint and
source-disjoint protocols preserve a reliable real-world evaluation
setting.

Regarding data release, all fully automatic pairs, generation code,
prompts, and configurations will be released. Redistributable manual
and semi-automatic pages will also be released together with their
character-level annotations. For archival materials restricted by
institutional authorization agreements, we will not redistribute the
original page images; instead, we will provide legally permissible
derived annotations, descriptive metadata, source identifiers, and
processing scripts. This release strategy balances reproducibility,
legal compliance, and the responsible use of culturally sensitive
historical materials.

\subsection{Inter-Annotator Agreement}
\label{app:annotation_agreement}

To assess annotation reliability, we measure inter-annotator agreement
on the manually collected real subset before expert adjudication.
Each page is independently annotated by two trained annotators with
native-language proficiency and prior experience in historical document
analysis. Since the annotation protocol contains fields with different
label structures, agreement is evaluated separately for character
transcription, visibility state, and degradation category.

For \textbf{character transcription}, agreement is measured using
character-level Cohen's $\kappa$ on aligned character instances,
achieving $\kappa=0.91$. For the \textbf{visibility state}
$v_i\in\{1,0,-1\}$, corresponding to \emph{clear},
\emph{partially degraded}, and \emph{severely degraded} characters,
Cohen's $\kappa$ reaches $0.86$. Since multiple degradation types may
co-occur within the same character or local region, agreement for the
\textbf{degradation category} $d_i$ is evaluated using Macro-F1,
achieving a score of $0.84$. These statistics are computed from the
initial independent annotations before expert adjudication, ensuring
that they reflect actual annotator consistency rather than post-hoc
correction.

After independent annotation, inconsistent cases are reviewed by a
senior expert according to a fixed annotation guideline covering
transcription conventions, visibility criteria, and degradation
definitions. In particular, the guideline distinguishes
\emph{partially degraded} characters, whose main structural evidence
remains locally observable, from \emph{severely degraded} characters,
whose identity cannot be reliably determined from local visual evidence
alone. The adjudicated annotations are used as the final supervision
for model training and benchmark evaluation.

\section{Methodology}
\label{app:methodology}
\subsection{Three-Stage Training Strategy}
\label{app:training_strategy}
Given the heterogeneous architecture of DDVR-Net, which couples a visual encoder with a lightweight language decoder, directly optimizing all modules jointly from scratch often leads to unstable training and suboptimal specialization across stages. To address this issue, we adopt a three-stage progressive training strategy that follows the intrinsic logic of the proposed framework: \textit{representation decoupling}, \textit{local perception learning}, and \textit{contextual restoration alignment}. This design ensures that the contextual decoder is only introduced after the visual branch has learned reliable glyph-aware and degradation-aware representations.

\subsubsection{Stage 1: Decoupled Visual Representation Learning}
In the first stage, we focus exclusively on Stage I and optimize the visual encoder together with the dual-branch projection modules. The objective of this stage is to disentangle character identity information from degradation-related corruption, so that the model can establish a stable glyph representation space and an explicit degradation representation space before any downstream recognition or restoration is performed. Concretely, the glyph branch is supervised by the contrastive glyph consistency objective, the degradation branch is optimized using degradation classification, and the two branches are further regularized by an orthogonality constraint to reduce representation entanglement. The optimization target in this stage is defined as:

\begin{equation}
\mathcal{L}_{Stage\_I} = \lambda_1 \mathcal{L}_{glyph} + \lambda_2 \mathcal{L}_{deg} + \lambda_3 \mathcal{L}_{ortho}
\end{equation}
where $\lambda_1, \lambda_2,$ and $\lambda_3$ are hyper-parameters balancing the contribution of each regularization term. Through this comprehensive decoupling, Stage I provides pure, noise-free character representations and explicit degradation context, establishing a robust foundation for the subsequent recognition and restoration stages.

\subsubsection{Stage 2: Local Perception and Reliability Modeling}
After the decoupled latent spaces become sufficiently stable, we introduce Stage II and train the local recognition and visibility prediction heads on top of the learned glyph and degradation features. At this stage, the model learns two complementary abilities: recognizing visually reliable characters from the glyph branch, and estimating the visibility state of each character from the degradation branch. Following the formulation in Section 4.2, severely degraded characters are excluded from the local recognition supervision, while all characters participate in visibility prediction. The training objective of this stage is:

\begin{equation}
\mathcal{L}_{Stage\_II} =\mathcal{L}_{Stage\_I}+ \lambda_{4}\mathcal{L}_{rec} + \lambda_{5}\mathcal{L}_{vis}
\end{equation}
where $\lambda_{4}$ and $\lambda_{5}$ are balancing coefficients that control the relative importance of the local recognition loss $\mathcal{L}_{rec}$ and the visibility prediction loss $\mathcal{L}_{vis}$, respectively. This stage equips the model with explicit local semantic perception and a reliability indicator that will later guide selective contextual restoration.

\subsubsection{Stage 3: Contextual Restoration Alignment}
In the third stage, we activate Stage III and introduce the character restoration decoder for degradation-conditioned contextual reasoning. To avoid disrupting the already learned visual representations, we retain the visual backbone as the primary perception component and only unfreeze the projection layer and the lightweight language decoder for alignment. Specifically, the decoder is fine-tuned with LoRA~\cite{hu2022lora} to improve parameter efficiency and training stability. During this stage, contextual restoration is only supervised on partially degraded or severely degraded characters, while clear characters continue to rely primarily on local visual recognition. The overall training objective becomes:
\begin{equation}
\mathcal{L}_{Stage\_III} = \mathcal{L}_{Stage\_II} + \lambda_{6}\mathcal{L}_{restore} + \lambda_{7}\mathcal{L}_{final}
\end{equation}
where $\lambda_{6}$ and $\lambda_{7}$ are balancing coefficients that control the relative contributions of the degradation-conditioned contextual restoration loss and the final fused recognition loss, respectively. This stage allows the model to gradually align visual evidence with contextual reasoning, while preventing the decoder from over-correcting visually clear characters.

\subsubsection{Training Discussion}
The above three-stage strategy follows a coarse-to-fine optimization path. Stage 1 establishes disentangled visual representations, Stage 2 learns reliable local perception and visibility estimation, and Stage 3 introduces context only when local evidence is insufficient. Compared with direct end-to-end optimization, this progressive design stabilizes multimodule training and better matches the degradation-aware restoration objective of historical documents.

\section{Experiments}
\label{app:experiments}
\subsection{Implementation and Efficiency}
\label{app:implementation_efficiency}
\subsubsection{Implementation Detail}
\label{app:implementation}

Our DDVR-Net framework is implemented using PyTorch. For the backbone architectures, we employ CogViT (0.4B) as the visual encoder to extract global spatial contexts and a lightweight large language model, Qwen2.5-0.5B-Instruct~\footnote{https://huggingface.co/Qwen/Qwen2.5-0.5B-Instruct}, as the character restoration decoder. To maintain parameter efficiency and training stability during the contextual restoration alignment stage, the language decoder is fine-tuned using LoRA~\cite{hu2022lora}. 

Following our three-stage progressive training strategy, the model is optimized using the AdamW~\cite{loshchilov2017decoupled} optimizer with $\beta_{1}=0.9$ and $\beta_{2}=0.999$. The base learning rate is set to $1 \times 10^{-4}$ with a cosine annealing schedule over 100 epochs. We train the network with a total batch size of 32 on 32 NVIDIA A100 GPUs. The input document images are processed at a normalized 300 dpi grayscale format. To alleviate the synthetic-to-real domain gap, our curriculum learning path begins with pretraining on the 150,000 fully automatic synthetic pairs, followed by adaptation on the 32,000 semi-automatic pairs, and concludes with fine-tuning on the manual real-world training split. Throughout the multi-task optimization, the loss balancing coefficients are empirically set to $\lambda_{1}=1.0$, $\lambda_{2}=0.5$, $\lambda_{3}=0.1$, $\lambda_{4}=1.0$, $\lambda_{5}=0.5$, $\lambda_{6}=0.8$, and $\lambda_{7}=1.0$.

\subsubsection{Efficiency and Computational Cost}
\label{app:efficiency}

We evaluate inference efficiency on a single NVIDIA A100 GPU using
batch size 1 and the same input preprocessing as in the main
experiments. Per-page latency is averaged over the complete real-world
test set after 20 warm-up iterations, with
\texttt{torch.cuda.synchronize()} called before and after each timed
forward pass. Peak GPU memory is measured using
\texttt{torch.cuda.max\_memory\_allocated()}. Training cost is reported
in GPU-hours, calculated as the number of GPUs multiplied by the
wall-clock training time. As shown in
Table~\ref{tab:efficiency}, DDVR-Net uses a compact visual encoder and
a Qwen2.5-0.5B-Instruct restoration decoder. Moreover, contextual restoration
is selectively weighted according to character visibility rather than
uniformly applied as a correction mechanism to all characters. This
design achieves a practical efficiency accuracy trade-off while
preserving the ability to recover severely degraded content.

\begin{table}[h]
\centering
\small
\caption{Efficiency and computational cost. Latency and memory are
measured on one NVIDIA A100 GPU with batch size 1. Training cost is
reported as the number of GPUs multiplied by wall-clock training time.}
\label{tab:efficiency}
\resizebox{\linewidth}{!}{
\begin{tabular}{lrrrrr}
\toprule
Model
& Total Params.
& Trainable Params.
& Latency (ms/page)
& Peak Memory (GB)
& Training Cost (GPU-hours) \\
\midrule
Local-only
& 407M & 407M & 286.4 & 7.8 & 4,096 \\
Uniform Context
& 913M & 415M & 612.7 & 12.6 & 5,888 \\
DDVR-Net
& 921M & 423M & 438.5 & 10.4 & 6,144 \\
\bottomrule
\end{tabular}}
\label{tab:dataset_statistics}
\end{table}

\subsection{Evaluation Metrics}
\label{app:evaluation_metrics}
Given the low-resource multilingual setting and the unified design of degradation perception, character recognition, and contextual restoration, we adopt a character-level evaluation protocol with task-specific metrics. Let $\mathcal{D}=\{(y_i,\hat{y}_i,v_i,\hat{v}_i,\mathbf{d}_i,\hat{\mathbf{d}}_i)\}_{i=1}^{N}$ denote the annotated character set, where $y_i$ and $\hat{y}_i$ are the ground-truth and predicted character labels, $v_i \in \{\texttt{clear},\texttt{partial},\texttt{severe}\}$ and $\hat{v}_i$ are the ground-truth and predicted visibility states, and $\mathbf{d}_i,\hat{\mathbf{d}}_i \in \{0,1\}^{K}$ are the ground-truth and predicted multi-hot degradation labels over $K$ degradation categories.

\subsubsection{Character Recognition and Restoration Metrics}
We first evaluate the final restored character prediction using character-level accuracy:
\begin{equation}
\mathrm{Acc}_{\mathrm{char}}
=
\frac{1}{N}\sum_{i=1}^{N}\mathbf{1}(\hat{y}_i = y_i)
\end{equation}
which corresponds to \textbf{Overall Character Accuracy}. To better measure restoration ability on difficult regions, we further define:
\begin{equation}
\mathrm{Acc}_{\mathrm{rest}}
=
\frac{1}{|\mathcal{I}_{\mathrm{deg}}|}
\sum_{i \in \mathcal{I}_{\mathrm{deg}}}\mathbf{1}(\hat{y}_i = y_i)
\end{equation}
where $\mathcal{I}_{\mathrm{deg}}=\{i \mid v_i \in \{\texttt{partial},\texttt{severe}\}\}$ denotes degraded characters. This is reported as \textbf{Restoration Accuracy}.

\begin{equation}
\mathrm{Acc}_{\mathrm{sev}}
=
\frac{1}{|\mathcal{I}_{\mathrm{sev}}|}
\sum_{i \in \mathcal{I}_{\mathrm{sev}}}\mathbf{1}(\hat{y}_i = y_i)
\end{equation}
where $\mathcal{I}_{\mathrm{sev}}=\{i \mid v_i = \texttt{severe}\}$ denotes severely degraded characters. This is reported as \textbf{Severely Degraded Accuracy}. In addition, we report the \textbf{Character Error Rate (CER)}:
\begin{equation}
\mathrm{CER}
=
\frac{\sum_{s \in \mathcal{S}} \mathrm{ED}(\hat{\mathbf{y}}^{(s)}, \mathbf{y}^{(s)})}
{\sum_{s \in \mathcal{S}} |\mathbf{y}^{(s)}|}
\end{equation}
where $\mathcal{S}$ is the set of text sequences, $\mathbf{y}^{(s)}$ and $\hat{\mathbf{y}}^{(s)}$ are the ground-truth and predicted character sequences, and $\mathrm{ED}(\cdot,\cdot)$ denotes the Levenshtein edit distance.

\subsubsection{Sequence-Level Metrics}
Since different scripts follow different reading orders, sequence construction is script-aware. For Tibetan, characters are serialized line by line from left to right; for Uyghur, line by line from right to left; and for Mongolian, column by column from top to bottom. Based on this rule, we define \textbf{Line-level Exact Match (LEM)} as:
\begin{equation}
\mathrm{LEM}
=
\frac{1}{|\mathcal{L}|}
\sum_{\ell \in \mathcal{L}}
\mathbf{1}(\hat{\mathbf{y}}^{(\ell)} = \mathbf{y}^{(\ell)})
\end{equation}
where $\mathcal{L}$ is the set of lines (or columns for Mongolian), and $\hat{\mathbf{y}}^{(\ell)}$ and $\mathbf{y}^{(\ell)}$ are the predicted and ground-truth ordered character sequences for unit $\ell$. Similarly, we report \textbf{Sequence Accuracy} at the page/document sequence level:
\begin{equation}
\mathrm{SeqAcc}
=
\frac{1}{|\mathcal{P}|}
\sum_{p \in \mathcal{P}}
\mathbf{1}(\hat{\mathbf{y}}^{(p)} = \mathbf{y}^{(p)})
\end{equation}
where $\mathcal{P}$ denotes the set of pages (or full text instances), and the sequence order follows the corresponding script-specific reading direction.

\subsubsection{Visibility Estimation Metrics}
To evaluate the explicit visibility modeling module, we report category-wise visibility accuracy:
\begin{equation}
\mathrm{Acc}_{c}
=
\frac{1}{|\mathcal{I}_{c}|}
\sum_{i \in \mathcal{I}_{c}}
\mathbf{1}(\hat{v}_i = v_i),
\quad
c \in \{\texttt{clear},\texttt{partial},\texttt{severe}\}
\end{equation}
where $\mathcal{I}_{c}=\{i \mid v_i = c\}$. We denote the three scores as \textbf{Acc\_clear}, \textbf{Acc\_partial}, and \textbf{Acc\_severe}, respectively.

\subsubsection{Degradation-Type Classification Metrics}
Because each character may be associated with multiple degradation types, degradation perception is evaluated as a multi-label classification task. We report both \textbf{Micro-F1} and \textbf{Macro-F1}. Let $\mathrm{TP}_k$, $\mathrm{FP}_k$, and $\mathrm{FN}_k$ be the true positives, false positives, and false negatives for degradation category $k$. Then:
\begin{equation}
\mathrm{Micro\mbox{-}F1}
=
\frac{2\sum_{k=1}^{K}\mathrm{TP}_k}
{2\sum_{k=1}^{K}\mathrm{TP}_k+\sum_{k=1}^{K}\mathrm{FP}_k+\sum_{k=1}^{K}\mathrm{FN}_k}
\end{equation}
and
\begin{equation}
\mathrm{Macro\mbox{-}F1}
=
\frac{1}{K}\sum_{k=1}^{K}
\frac{2\mathrm{TP}_k}
{2\mathrm{TP}_k+\mathrm{FP}_k+\mathrm{FN}_k}
\end{equation}
Unless otherwise specified, all metrics are computed on the test set at the character level, while sequence-level metrics additionally respect the native reading order of each script.

\subsection{Unified Evaluation Protocol for General-Purpose Baselines}
\label{app:baseline_protocol}

We evaluate all open and closed-source general-purpose vision-language models under a unified end-to-end setting. Models receive only the full document image without ground-truth boxes, lexicons, or external retrieval and a standardized prompt. No model-specific engineering is applied. 

\textbf{Unified instruction template} To enforce a consistent output format and script-specific reading order, all baselines are queried with the identical instruction:

\begin{lstlisting}[breaklines=true, basicstyle=\ttfamily\small, columns=fullflexible, keepspaces=true]
You are given one full-page image of a historical document written in
one of the following scripts: Tibetan, Mongolian, or Uyghur.

Your task is to perform end-to-end character-level document understanding
on this page.

Important requirements:
1. Read the page in the correct native reading order:
   - Tibetan: line by line, left to right.
   - Uyghur: line by line, right to left.
   - Mongolian: column by column, top to bottom.
2. Output an ordered sequence of character tokens for the whole page.
3. For each character token, predict:
   - "char": the recognized or restored character,
   - "visibility": one label from {"clear", "partial", "severe"},
   - "degradation": zero, one, or multiple labels from:
     {"character$_missing", "stroke_fading", "structural_breakage",
      "local_blur", "ink_erosion", "local_occlusion",
      "cracks", "wormholes", "stains", "folds", "paper_damage"$}$.
4. If the local visual evidence is clear, preserve the visually observed
   character.
5. If the local visual evidence is partially damaged or severely corrupted,
   restore the character using both local evidence and surrounding textual
   context.
6. Do not output explanations, reasoning, markdown, or extra text.
7. Output JSON only, following exactly this schema:$

{
  "script": "Tibetan or Mongolian or Uyghur",
  "reading$_order": "line_ltr / line_rtl / column_ttb",
  "tokens": [
    {
      "char": "...",
      "visibility": "clear | partial | severe",
      "degradation": ["label1", "label2"]
    }
  ]
$}$
$\end{lstlisting}

\textbf{Decoding and output parsing.} We use greedy or temperature-0 decoding whenever available. To mitigate residual API stochasticity, each model is queried eight times, and we report the averaged metrics. Responses are parsed by extracting the largest valid JSON block, with minor formatting errors automatically corrected by a deterministic script. Unparsable responses are treated as failed predictions (empty sequences).

\subsection{Main Results}
\label{app:main_results}
To better understand the contribution of different data sources, we further analyze performance on each test subset separately. In general, we expect a consistent difficulty order of \textit{fully automatic} $<$ \textit{semi-automatic} $<$ \textit{manual real-world}, i.e., \textbf{Fully-auto only} should yield the best results, followed by \textbf{Semi-auto only}, while \textbf{Manual real only} should remain the most challenging. This trend is consistent with the dataset statistics in Table~\ref{tab:dataset_statistics}: the manual subset contains substantially more damaged characters per image than the semi-automatic and fully automatic subsets across all three languages, indicating heavier degradation and a larger real-world distribution gap. On the manual real-world test set, the full curriculum model achieves 86.47/74.29/61.54/9.83/36.75 on Tibetan, 84.35/71.18/58.42/11.23/32.49 on Mongolian, and 85.34/72.87/59.61/
10.45/34.29 on Uyghur in terms of Acc$_{char}$/Acc$_{rest}$/Acc$_{sev}$/CER/
SeqAcc, respectively. These results indicate that although manual real-world pages remain the hardest split, the proposed framework still maintains stable recognition and restoration performance under authentic degradation.

\begin{table}[t]
\centering
\caption{Performance of the full model on the manual real-world test subset. Higher is better except CER.}
\label{tab:manual_only_test}
\resizebox{\columnwidth}{!}{
\begin{tabular}{lccccc}
\toprule
\textbf{Lang.} & $\mathbf{Acc_{char}}$ & $\mathbf{Acc_{rest}}$ & $\mathbf{Acc_{sev}}$ & $\mathbf{CER}\downarrow$ & $\mathbf{SeqAcc}$ \\
\midrule
Tibetan   & 86.47 & 74.29 & 61.54 & 9.83  & 36.75 \\
Mongolian & 84.35 & 71.18 & 58.42 & 11.23 & 32.49 \\
Uyghur    & 85.34 & 72.87 & 59.61 & 10.45 & 34.29 \\
\bottomrule
\end{tabular}}
\end{table}

\begin{table}[t]
\centering
\caption{Performance of the full model on the semi-automatic test subset. Higher is better except CER.}
\label{tab:semi_auto_only_test}
\resizebox{\columnwidth}{!}{
\begin{tabular}{lccccc}
\toprule
\textbf{Lang.} & $\mathbf{Acc_{char}}$ & $\mathbf{Acc_{rest}}$ & $\mathbf{Acc_{sev}}$ & $\mathbf{CER}\downarrow$ & $\mathbf{SeqAcc}$ \\
\midrule
Tibetan   & 89.12 & 78.96 & 67.84 & 8.21 & 42.63 \\
Mongolian & 87.45 & 75.83 & 64.29 & 9.56 & 38.47 \\
Uyghur    & 88.31 & 77.14 & 65.72 & 8.88 & 40.15 \\
\bottomrule
\end{tabular}}
\end{table}

As shown in Table \ref{tab:semi_auto_only_test}, compared with the manual real-world test subset, the semi-automatic subset yields consistently stronger results across all three languages. This trend is expected because the semi-automatic data preserves more regular layout and cleaner structural evidence, while still retaining degradation patterns derived from real historical pages. In particular, the gains on Acc$_{rest}$ and Acc$_{sev}$ indicate that the proposed visibility-guided restoration remains effective when degradation is present but less destructive than in fully authentic real-world pages. Meanwhile, the lower CER and higher SeqAcc suggest that the model benefits from the relatively reduced domain complexity of the semi-automatic subset while still maintaining robust restoration behavior.

\begin{table}[t]
\centering
\caption{Performance of the full model on the fully automatic test subset. Higher is better except CER.}
\label{tab:fully_auto_only_test}
\resizebox{\columnwidth}{!}{
\begin{tabular}{lccccc}
\toprule
\textbf{Lang.} & $\mathbf{Acc_{char}}$ & $\mathbf{Acc_{rest}}$ & $\mathbf{Acc_{sev}}$ & $\mathbf{CER}\downarrow$ & $\mathbf{SeqAcc}$ \\
\midrule
Tibetan   & 91.36 & 82.74 & 72.18 & 6.94 & 47.58 \\
Mongolian & 89.84 & 79.61 & 68.43 & 8.07 & 43.26 \\
Uyghur    & 90.52 & 80.88 & 69.95 & 7.42 & 45.11 \\
\bottomrule
\end{tabular}}
\end{table}

As shown in Table \ref{tab:fully_auto_only_test}, the fully automatic test subset yields the strongest performance among the three evaluation splits. This trend is consistent with the data construction process, since the fully automatic subset provides cleaner layout regularity and more controlled degradation patterns than the manual real-world and semi-automatic subsets. Across all three languages, the further gains in Acc$_{rest}$ and Acc$_{sev}$ show that the proposed framework is particularly effective when degradation remains challenging but follows a more stable synthetic distribution. Meanwhile, the lowest CER and highest SeqAcc suggest that the model can better exploit the relatively consistent visual and textual structure of the fully automatic subset.

\subsection{Ablation Study}
\label{app:ablation}
\subsubsection{Fair Baselines under the Same Backbone and Data}
\label{app:fair_baselines}

\begin{table*}[h]
\centering
\caption{Fair comparison under the same backbone and training data across different languages. Higher is better except CER.}
\label{tab:fair_baselines_lang}
\setlength{\tabcolsep}{4pt}
% 2. 极限压缩行高（从 0.9 降到 0.75）
\renewcommand{\arraystretch}{1} 
% 3. 消除 toprule, midrule, bottomrule 默认的上下留白
\setlength{\aboverulesep}{0pt}
\setlength{\belowrulesep}{0pt}
\resizebox{\linewidth}{!}{
\begin{tabular}{c l ccccc}
\toprule
\multirow{2}{*}{\textbf{Lang.}} 
& \multirow{2}{*}{\textbf{Training Paradigm}} 
& \multicolumn{4}{c}{\textbf{Character Recognition \& Restoration}} 
& \textbf{Sequence-Level} \\
\cmidrule(lr){3-6} \cmidrule(lr){7-7}
& 
& $\mathrm{Acc}_{\mathrm{char}}$ 
& $\mathrm{Acc}_{\mathrm{rest}}$ 
& $\mathrm{Acc}_{\mathrm{sev}}$ 
& $\mathrm{CER} \downarrow$ 
& $\mathrm{SeqAcc}$ \\
\midrule

\multirow{5}{*}{\rotatebox[origin=c]{90}{\textbf{Tibetan}}}
& Full DDVR-Net & \textbf{86.47} & \textbf{74.29} & \textbf{61.54} & \textbf{9.83} & \textbf{36.75} \\
& Confidence-based Correction & 85.10 & 73.60 & 62.10 & 10.90 & 32.60 \\
& Post-hoc Correction & 84.80 & 72.90 & 59.80 & 11.20 & 31.80 \\
& Uniform Context & 84.20 & 72.80 & 60.90 & 11.70 & 29.40 \\
& Local-only & 83.90 & 68.90 & 54.60 & 12.80 & 30.70 \\
\midrule

\multirow{5}{*}{\rotatebox[origin=c]{90}{\textbf{Mongolian}}}
& Full DDVR-Net & \textbf{84.35} & \textbf{71.18} & \textbf{58.42} & \textbf{11.23} & \textbf{32.49} \\
& Confidence-based Correction & 82.40 & 70.40 & 58.60 & 12.20 & 28.70 \\
& Post-hoc Correction & 82.00 & 69.90 & 56.90 & 12.60 & 27.90 \\
& Uniform Context & 81.20 & 69.70 & 57.30 & 13.00 & 25.80 \\
& Local-only & 80.70 & 65.40 & 50.80 & 14.10 & 26.90 \\
\midrule

\multirow{5}{*}{\rotatebox[origin=c]{90}{\textbf{Uyghur}}}
& Full DDVR-Net & \textbf{85.34} & \textbf{72.87} & \textbf{59.61} & \textbf{10.45} & \textbf{34.29} \\
& Confidence-based Correction & 83.70 & 72.20 & 60.30 & 11.50 & 30.40 \\
& Post-hoc Correction & 83.20 & 71.60 & 58.40 & 11.80 & 29.60 \\
& Uniform Context & 82.50 & 71.10 & 58.90 & 12.30 & 27.50 \\
& Local-only & 81.90 & 66.80 & 52.70 & 13.40 & 28.60 \\
\bottomrule
\end{tabular}}
\end{table*}

To further verify that the gains of DDVR-Net come from the proposed selective restoration design rather than stronger backbone capacity or additional data, we compare with several controlled baselines under the same visual encoder, contextual decoder, training data, and training schedule. Specifically, \textbf{Uniform Context} applies contextual restoration to all characters without selective gating; \textbf{Confidence-based Correction} triggers restoration according to the local recognition confidence; \textbf{Post-hoc Correction} first performs local recognition and then applies a separate correction stage without joint token-level fusion; and \textbf{Local-only} removes the contextual decoder entirely.

Table~\ref{tab:fair_baselines_lang} shows that the improvement of DDVR-Net does not simply come from adding a contextual decoder. Across all three languages, \textbf{Uniform Context} consistently improves degraded-character recovery over \textbf{Local-only}, particularly on $\mathrm{Acc}_{rest}$ and $\mathrm{Acc}_{sev}$, but yields lower $\mathrm{SeqAcc}$, indicating that indiscriminate contextual correction tends to over-correct visually reliable characters. \textbf{Confidence-based Correction} is more competitive and achieves better balance than \textbf{Uniform Context}, suggesting that selective triggering is necessary; however, it still remains below the full model on nearly all metrics, which indicates that local prediction confidence alone is not a sufficiently reliable proxy for degradation visibility. \textbf{Post-hoc Correction} also improves over \textbf{Local-only}, but is consistently weaker than DDVR-Net, showing that applying correction after local recognition is less effective than jointly learned visibility-guided fusion. Overall, these results across Tibetan, Mongolian, and Uyghur support that the key advantage of DDVR-Net lies in \emph{when} and \emph{how} contextual restoration is activated, rather than merely in using the same backbone with an additional decoder.

\subsubsection{Effect of Degradation-Aware Representation Decoupling}
\label{app:representation_decoupling}

Stage I of our framework introduces a Degradation-Aware Representation Learning paradigm. To validate that our performance gains stem from the genuine disentanglement of intrinsic glyph information and extrinsic environmental noise, rather than merely from increased model capacity, we perform a comprehensive ablation study on the decoupling mechanism. As shown in Table \ref{tab:ablation_decoupling}, we compare the Full DDVR-Net against four variants: 
(1) \textbf{w/o Decoupling}: removes the dual-branch design, feeding RoI features directly into the prediction heads; 
(2) \textbf{w/o Ortho}: removes the orthogonality loss ($\mathcal{L}_{ortho}$); 
(3) \textbf{w/o Glyph Consistency}: removes the Supervised Contrastive Learning ($\mathcal{L}_{glyph}$) on the glyph branch; 
(4) \textbf{w/o Deg Cls}: removes the multi-label degradation classification ($\mathcal{L}_{deg}$).

The results demonstrate that w/o Decoupling suffers the most significant degradation across all metrics, particularly in $\mathrm{Acc}_{\mathrm{sev}}$, $\mathrm{Acc}_{\mathrm{rest}}$, and CER. Without explicit disentanglement, the model easily misinterprets local deterioration patterns (e.g., cracks, stains, ink erosion) as genuine glyph structures, severely deteriorating performance on partially and severely degraded samples. Furthermore, analyzing the specific constraints reveals their distinct roles. Removing the orthogonality penalty (w/o Ortho) leads to a noticeable performance drop, proving its necessity as a stabilizer to prevent information leakage between the two latent subspaces. The w/o Glyph Consistency variant primarily harms character recognition accuracy, as the model loses the cross-degradation consistency required to cluster identical characters suffering from different damage modes. Lastly, w/o Deg Cls severely compromises visibility estimation and degradation F1 scores, which indirectly cripples the downstream selective restoration process. Ultimately, these findings confirm that the Stage I decoupling is not a dispensable auxiliary module, but the fundamental prerequisite that enables accurate visibility estimation and reliable, degradation-conditioned contextual restoration.
%证明 Stage I 的必要性，突出 解耦表示是后续任务的基础。
\begin{table*}[h]
\centering
\caption{Effect of Degradation-Aware Representation Decoupling}
\label{tab:ablation_decoupling}
\setlength{\tabcolsep}{4pt}
% 极限压缩行高至 0.6
\renewcommand{\arraystretch}{1}
% 消除横线默认的上下留白，确保极限压缩生效
\setlength{\aboverulesep}{0pt}
\setlength{\belowrulesep}{0pt}

\resizebox{\linewidth}{!}{
\footnotesize % 配合极小行高，整体字号也调小
\begin{tabular}{c|l|cccc|cc|cc}
\toprule
\multirow{2}{*}{\textbf{Lang.}} 
& \multirow{2}{*}{\textbf{Models}} 
& \multicolumn{4}{c|}{\textbf{Character Recognition \& Restoration}}  
& \multicolumn{2}{c|}{\textbf{Visibility Estimation}} 
& \multicolumn{2}{c}{\textbf{Degradation-Type Classification}} \\
\cmidrule(lr){3-6}\cmidrule(lr){7-8} \cmidrule(lr){9-10}
& 
& $\mathrm{Acc}_{\mathrm{char}}$ 
& $\mathrm{Acc}_{\mathrm{rest}}$ 
& $\mathrm{Acc}_{\mathrm{sev}}$ 
& $\mathrm{CER} \downarrow$ 
& $\mathrm{Acc}_{\mathrm{partial}}$ 
& $\mathrm{Acc}_{\mathrm{severe}}$ 
& $\mathrm{Micro\mbox{-}F1}$ 
& $\mathrm{Macro\mbox{-}F1}$ \\
\midrule

% Tibetan
\multirow{5}{*}{\rotatebox[origin=c]{90}{\scriptsize \textbf{Tibetan}}} 
& \textbf{Full DDVR-Net}    & \textbf{86.47} & \textbf{74.29} & \textbf{61.54} & \textbf{9.83}  & \textbf{78.46} & \textbf{69.18} & \textbf{81.24} & \textbf{79.67} \\
& w/o Decoupling           & 78.50 & 58.20 & 42.30 & 15.50 & 62.10 & 51.40 & 65.20 & 63.10 \\
& w/o Ortho                & 84.10 & 70.50 & 57.20 & 10.90 & 75.30 & 66.10 & 77.80 & 76.20 \\
& w/o Glyph Consistency    & 82.30 & 68.40 & 55.10 & 11.80 & 76.50 & 67.40 & 79.10 & 78.00 \\
& w/o Deg Cls              & 83.50 & 64.20 & 48.60 & 13.20 & 58.40 & 46.20 & - & - \\ \midrule

% Mongolian
\multirow{5}{*}{\rotatebox[origin=c]{90}{\scriptsize \textbf{Mongolian}}} 
& \textbf{Full DDVR-Net}    & \textbf{84.35} & \textbf{71.18} & \textbf{58.42} & \textbf{11.23} & \textbf{75.84} & \textbf{66.27} & \textbf{78.53} & \textbf{76.92} \\
& w/o Decoupling           & 76.21 & 54.36 & 39.15 & 17.48 & 59.42 & 48.11 & 62.15 & 60.34 \\
& w/o Ortho                & 82.14 & 67.85 & 54.62 & 12.35 & 72.16 & 63.45 & 75.28 & 73.81 \\
& w/o Glyph Consistency    & 80.52 & 65.41 & 52.18 & 13.56 & 73.85 & 64.12 & 76.42 & 74.95 \\
& w/o Deg Cls              & 81.36 & 61.25 & 46.58 & 14.82 & 55.34 & 44.27 & - & - \\ \midrule

% Uyghur
\multirow{5}{*}{\rotatebox[origin=c]{90}{\scriptsize \textbf{Uyghur}}} 
& \textbf{Full DDVR-Net}    & \textbf{85.34} & \textbf{72.87} & \textbf{59.61} & \textbf{10.45} & \textbf{76.92} & \textbf{67.35} & \textbf{79.46} & \textbf{77.83} \\
& w/o Decoupling           & 77.45 & 56.12 & 40.85 & 16.21 & 60.58 & 49.62 & 63.74 & 61.92 \\
& w/o Ortho                & 83.15 & 69.45 & 55.82 & 11.64 & 73.54 & 64.51 & 76.12 & 74.85 \\
& w/o Glyph Consistency    & 81.28 & 67.14 & 53.46 & 12.78 & 75.12 & 65.84 & 77.65 & 75.91 \\
& w/o Deg Cls              & 82.41 & 63.15 & 47.38 & 14.05 & 57.26 & 45.18 & - & - \\

\bottomrule
\end{tabular}
}
\end{table*}

\subsubsection{Effect of Visibility-Guided Selective Restoration}
\label{app:visibility_restoration}

To validate the necessity of applying contextual reasoning selectively rather than universally, we evaluate the visibility-guided restoration mechanism. As shown in Table \ref{tab:ablation_visibility}, we compare Full DDVR-Net against four variants: (1) \textbf{Uniform Context for All}, which applies the language decoder to all characters without the degradation mask; (2) \textbf{w/o Visibility Head}, which removes explicit visibility prediction; (3) \textbf{Hard Switch}, which replaces the soft visibility-guided gate ($\alpha_i$) with a binary threshold; and (4) \textbf{Local-only}, which completely removes the Stage III contextual decoder.

\begin{figure*}[h]
    \centering
    \begin{minipage}{0.48\textwidth}
        \centering
        \includegraphics[width=\linewidth, height=6cm]{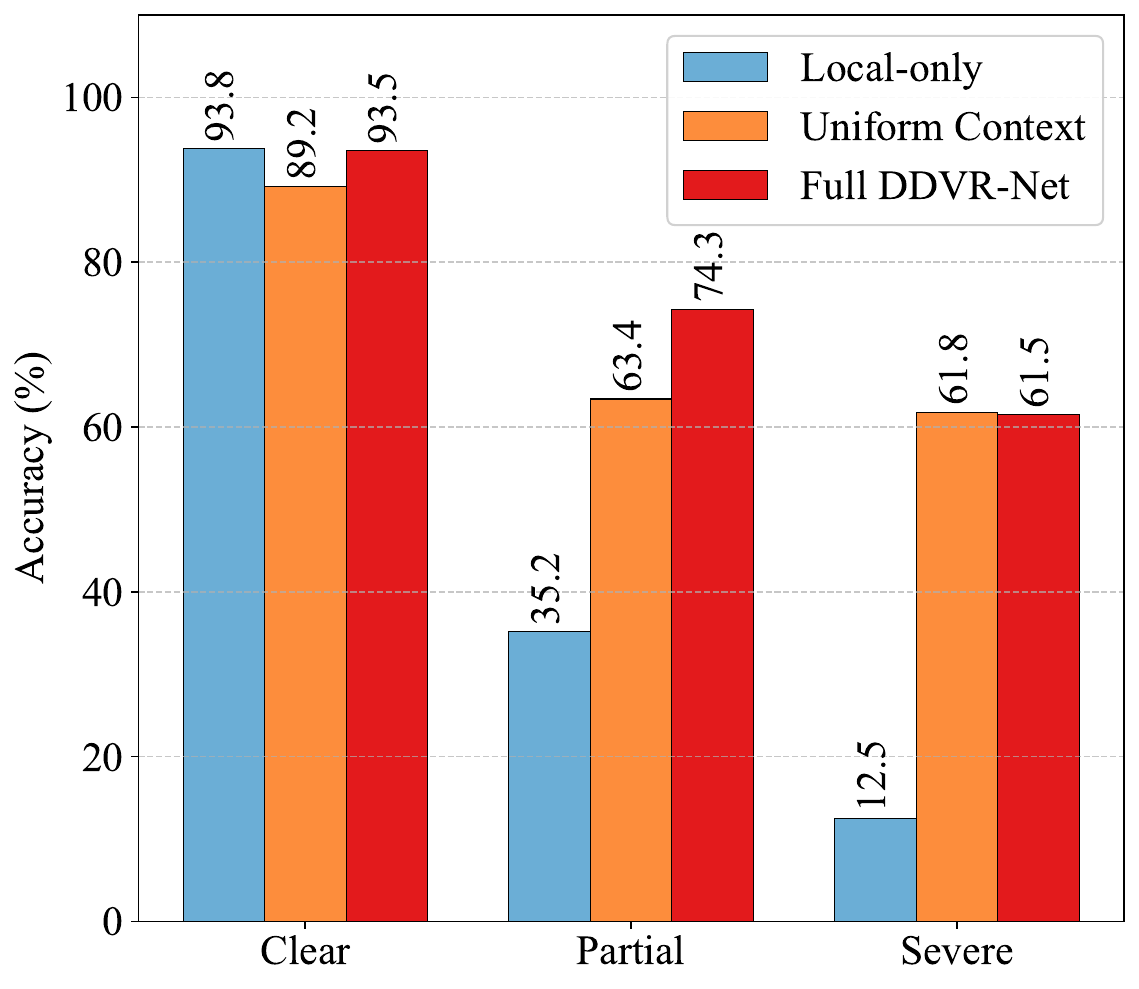}
        \centerline{\tiny{(a) Performance Stratified by Visibility}}
    \end{minipage}
    \hfill
    \begin{minipage}{0.48\textwidth}
        \centering
        \includegraphics[width=\linewidth, height=6cm]{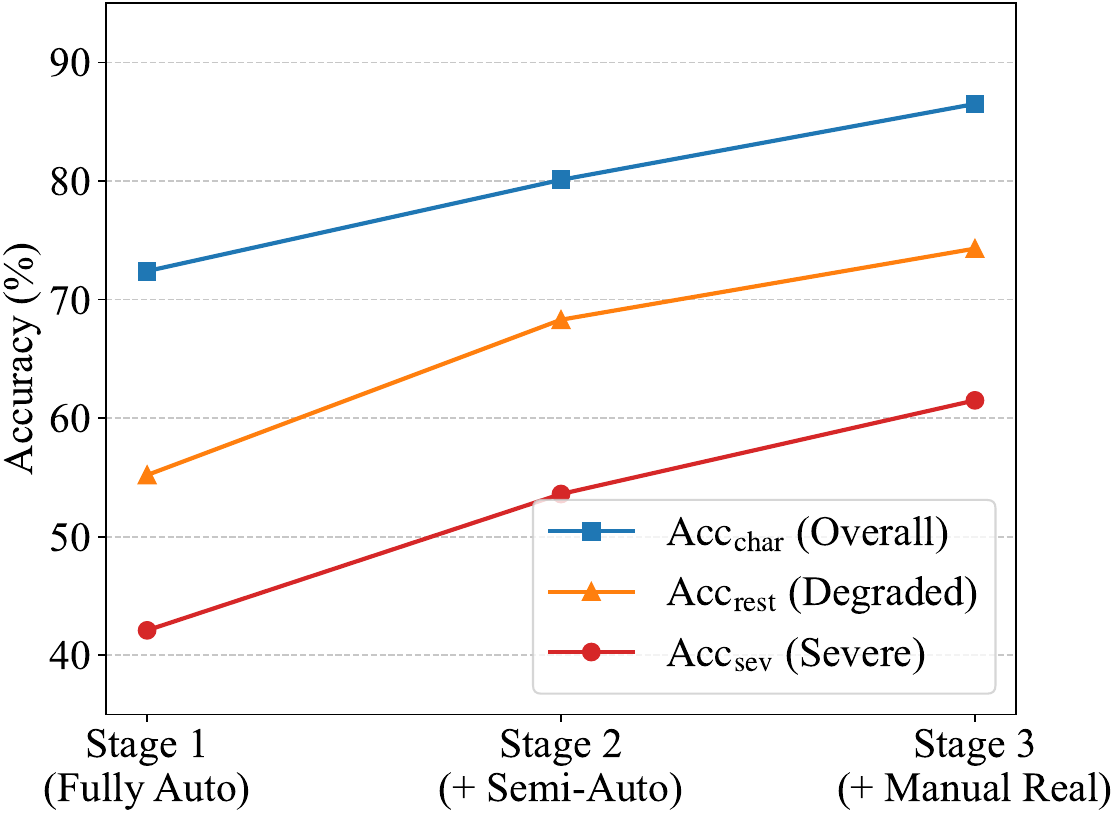}
        \centerline{\tiny{(b) Effect of Progressive Curriculum}}
    \end{minipage}
     \vspace{-4mm}
    \caption{Detailed analysis of DDVR-Net components. (a) Stratifying performance by visibility explicitly demonstrates the trade-off resolved by our visibility-guided selective restoration: it preserves high accuracy on Clear characters while fully recovering Severe characters, and optimally fuses signals for Partial characters. (b) Progressive integration of heterogeneous datasets continuously improves performance, especially on severe degradation.}
     \vspace{-5mm}
    \label{fig:ablation_analysis}
\end{figure*}

The results, as further visualized by the visibility-stratified performance in Figure \ref{fig:ablation_analysis} (a), reveal a clear trade-off inherent in historical document restoration. The Local-only variant achieves high accuracy on clear characters ($\mathrm{Acc}_{\mathrm{clear-char}}$) but fails catastrophically on severely degraded regions ($\mathrm{Acc}_{\mathrm{sev}}$), proving that contextual reasoning is indispensable when visual evidence is corrupted. Conversely, Uniform Context for All successfully recovers degraded text but suffers a noticeable drop in $\mathrm{Acc}_{\mathrm{clear-char}}$ and sequence-level metrics (LEM, SeqAcc). This confirms that blindly forcing a language model to process visually unambiguous characters leads to over-correction, where the model "guesses" based on context rather than "reading" the clear visual evidence.

Furthermore, the Hard Switch variant performs worse than our soft gating mechanism, indicating that visibility in real historical documents is not strictly binary; partially degraded characters benefit from a continuous, learned fusion of local visual cues and contextual priors. Ultimately, these findings demonstrate that DDVR-Net's superior performance does not merely stem from appending a language decoder, but rather from learning exactly when to activate it and how much to trust it, acting as a targeted fallback mechanism rather than a universal correction module.
\begin{table*}[h]
\centering
\caption{Effect of Visibility-Guided Selective Restoration}
\label{tab:ablation_visibility}
\setlength{\tabcolsep}{4pt}
% 按照您的要求，极限压缩行高至 0.6
\renewcommand{\arraystretch}{1} 
% 必须消除横线默认的上下留白，否则达不到极限压缩的效果
\setlength{\aboverulesep}{0pt}
\setlength{\belowrulesep}{0pt}

\resizebox{\linewidth}{!}{
\footnotesize % 配合极小行高，整体字号也调小
\begin{tabular}{c|l|cccc|cc|c}
\toprule
\multirow{2}{*}{\textbf{Lang.}} 
& \multirow{2}{*}{\textbf{Models}} 
& \multicolumn{4}{c|}{\textbf{Character Recognition \& Restoration}}  
& \multicolumn{2}{c|}{\textbf{Sequence-Level}} 
& \multicolumn{1}{c}{\textbf{Clear Chars}} \\
\cmidrule(lr){3-6}\cmidrule(lr){7-8} \cmidrule(lr){9-9}
& 
& $\mathrm{Acc}_{\mathrm{char}}$ 
& $\mathrm{Acc}_{\mathrm{rest}}$ 
& $\mathrm{Acc}_{\mathrm{sev}}$ 
& $\mathrm{CER} \downarrow$ 
& $\mathrm{LEM}$ 
& $\mathrm{SeqAcc}$ 
& $\mathrm{Acc_{clear-char}}$  \\
\midrule

% Tibetan
\multirow{5}{*}{\rotatebox[origin=c]{90}{\scriptsize \textbf{Tibetan}}} 
& \textbf{Full DDVR-Net}            & \textbf{86.47} & 74.29 & 61.54 & \textbf{9.83}  & \textbf{45.62} & \textbf{36.75} & 93.50 \\ 
& Uniform Context for All          & 84.10 & \textbf{74.50} & \textbf{61.80} & 11.50 & 42.10 & 33.50 & 89.20 \\
& w/o Visibility Head              & 83.20 & 68.50 & 55.40 & 12.30 & 40.50 & 31.80 & 91.00 \\
& Hard Switch instead of Soft Gate & 85.10 & 71.80 & 58.90 & 10.60 & 43.80 & 34.90 & 92.80 \\
& Local-only                       & 75.40 & 35.20 & 12.50 & 22.40 & 25.60 & 18.50 & \textbf{93.80} \\ \midrule

% Mongolian
\multirow{5}{*}{\rotatebox[origin=c]{90}{\scriptsize \textbf{Mongolian}}} 
& \textbf{Full DDVR-Net}            & \textbf{84.35} & 71.18 & 58.42 & \textbf{11.23} & \textbf{41.76} & \textbf{32.49} & 92.10 \\ 
& Uniform Context for All          & 82.50 & \textbf{71.50} & \textbf{58.60} & 12.80 & 38.50 & 29.80 & 87.50 \\
& w/o Visibility Head              & 81.40 & 66.30 & 53.20 & 13.90 & 37.20 & 28.50 & 90.10 \\
& Hard Switch instead of Soft Gate & 83.20 & 69.10 & 56.30 & 12.10 & 40.10 & 31.00 & 91.50 \\
& Local-only                       & 73.80 & 33.50 & 10.80 & 24.10 & 22.40 & 15.20 & \textbf{92.40} \\ \midrule

% Uyghur
\multirow{5}{*}{\rotatebox[origin=c]{90}{\scriptsize \textbf{Uyghur}}} 
& \textbf{Full DDVR-Net}            & \textbf{85.34} & 72.87 & 59.61 & \textbf{10.45} & \textbf{43.58} & \textbf{34.29} & 93.10 \\ 
& Uniform Context for All          & 83.20 & \textbf{73.10} & \textbf{59.80} & 12.10 & 40.20 & 31.50 & 88.40 \\
& w/o Visibility Head              & 82.10 & 67.50 & 54.80 & 13.10 & 38.50 & 30.10 & 90.80 \\
& Hard Switch instead of Soft Gate & 84.10 & 70.40 & 57.50 & 11.20 & 41.80 & 32.80 & 92.20 \\
& Local-only                       & 74.50 & 34.10 & 11.50 & 23.50 & 24.10 & 16.80 & \textbf{93.30} \\ 
\bottomrule
\end{tabular}
}
\end{table*}

\subsubsection{Data Composition and Progressive Curriculum}
\label{app:data_curriculum}
\label{sec:ablation_curriculum}

Our framework leverages a newly constructed dataset comprising 187,000
document pairs from three heterogeneous sources: fully automatic,
semi-automatic, and manual real-world data. We distinguish the effect
of data-source composition from that of curriculum order by evaluating
all variants with the same architecture exclusively on the
document-disjoint manual real-world test sets. Fully automatic and
semi-automatic data provide scalable degradation and layout priors but
retain a measurable synthetic-to-real domain gap, whereas manual-only
training avoids synthetic artifacts but is limited by its small scale.
We therefore adopt a progressive
Auto$\rightarrow$Semi$\rightarrow$Manual curriculum that performs
large-scale pre-training on fully automatic data, intermediate
adaptation on semi-automatic data, and final calibration on manual
real-world data. As visualized in
Figure~\ref{fig:ablation_analysis}(b), this stage-wise integration
continuously improves performance, particularly for severely degraded
characters measured by $\mathrm{Acc}_{\mathrm{sev}}$.

To validate both data composition and curriculum order, we compare five
training paradigms: \textbf{Manual Only}, \textbf{w/o Manual Fine-tune},
\textbf{Mixed Training}, \textbf{Reverse Curriculum}, and the proposed
\textbf{Full Curriculum}. As shown in
Table~\ref{tab:ablation_curriculum}, \textbf{Manual Only} yields poor
restoration performance, indicating that 5,000 real pairs are
insufficient for learning diverse degradation priors. Conversely,
\textbf{w/o Manual Fine-tune} retains a pronounced synthetic-to-real
gap and struggles with authentic physical corruption. Simply combining
all subsets through \textbf{Mixed Training} is also suboptimal because
the substantially larger synthetic subset dominates the training
distribution. Moreover, \textbf{Reverse Curriculum}
(Manual$\rightarrow$Semi$\rightarrow$Auto) degrades performance by
shifting the calibrated representations back toward the synthetic
domain. In contrast, the proposed \textbf{Full Curriculum} consistently
achieves the strongest results, showing that the improvement depends
not merely on data volume but on a structured progression from
synthetic scale to authentic supervision. Specifically, the fully
automatic subset provides large-scale visual and textual priors, the
semi-automatic subset introduces historically plausible layouts and
degradation patterns, and the manual subset performs final real-world
distribution calibration. This progressive adaptation enables robust
degradation-aware recognition and restoration under realistic
conditions.

\begin{table*}[h]
\centering
\caption{Effect of Progressive Curriculum over Heterogeneous Data Sources (Evaluated on Manual Real-World Test Set)}
\label{tab:ablation_curriculum}
% 1. 缩小列间距
\setlength{\tabcolsep}{4pt}
% 2. 极限压缩行高（从 0.9 降到 0.75）
\renewcommand{\arraystretch}{1} 
% 3. 消除 toprule, midrule, bottomrule 默认的上下留白
\setlength{\aboverulesep}{0pt}
\setlength{\belowrulesep}{0pt}

\resizebox{\linewidth}{!}{
% 4. 整体使用更小的字号，进一步减小基础行距
\footnotesize 
\begin{tabular}{c|l|ccccc}
\toprule
\multirow{2}{*}{\textbf{Lang.}} 
& \multirow{2}{*}{\textbf{Training Paradigm}} 
& \multicolumn{4}{c}{\textbf{Character Recognition \& Restoration}} 
& \textbf{Sequence-Level} \\
\cmidrule(lr){3-6} \cmidrule(lr){7-7}
& 
& $\mathrm{Acc}_{\mathrm{char}}$ 
& $\mathrm{Acc}_{\mathrm{rest}}$ 
& $\mathrm{Acc}_{\mathrm{sev}}$ 
& $\mathrm{CER} \downarrow$ 
& $\mathrm{SeqAcc}$ \\
\midrule

% 藏语部分
\multirow{5}{*}{\rotatebox[origin=c]{90}{\scriptsize \textbf{Tibetan}}}
& \textbf{Full Curriculum (Auto $\rightarrow$ Semi $\rightarrow$ Manual)} & \textbf{86.47} & \textbf{74.29} & \textbf{61.54} & \textbf{9.83}  & \textbf{36.75} \\
& Mixed Training (Random Shuffle)                                         & 83.15 & 69.80 & 56.20 & 12.10 & 32.40 \\
& Reverse Curriculum (Manual $\rightarrow$ Semi $\rightarrow$ Auto)       & 80.40 & 65.30 & 52.10 & 14.50 & 28.50 \\
& w/o Manual Fine-tune (Auto $\rightarrow$ Semi)                          & 81.20 & 66.80 & 54.50 & 13.80 & 29.80 \\
& Manual Only                                                             & 76.50 & 58.40 & 41.20 & 18.20 & 22.10 \\ \midrule

% 蒙古语部分 
\multirow{5}{*}{\rotatebox[origin=c]{90}{\scriptsize \textbf{Mongolian}}}
& \textbf{Full Curriculum (Auto $\rightarrow$ Semi $\rightarrow$ Manual)} & \textbf{84.35} & \textbf{71.18} & \textbf{58.42} & \textbf{11.23} & \textbf{32.49} \\
& Mixed Training (Random Shuffle)                                         & 80.90 & 66.50 & 53.10 & 13.50 & 28.10 \\
& Reverse Curriculum (Manual $\rightarrow$ Semi $\rightarrow$ Auto)       & 78.20 & 62.10 & 49.30 & 16.10 & 24.30 \\
& w/o Manual Fine-tune (Auto $\rightarrow$ Semi)                          & 79.10 & 63.80 & 51.20 & 15.40 & 25.80 \\
& Manual Only                                                             & 74.30 & 55.20 & 38.50 & 20.50 & 18.40 \\ \midrule

% 维吾尔语部分 
\multirow{5}{*}{\rotatebox[origin=c]{90}{\scriptsize \textbf{Uyghur}}}
& \textbf{Full Curriculum (Auto $\rightarrow$ Semi $\rightarrow$ Manual)} & \textbf{85.34} & \textbf{72.87} & \textbf{59.61} & \textbf{10.45} & \textbf{34.29} \\
& Mixed Training (Random Shuffle)                                         & 82.10 & 68.20 & 54.80 & 12.80 & 30.10 \\
& Reverse Curriculum (Manual $\rightarrow$ Semi $\rightarrow$ Auto)       & 79.50 & 64.10 & 50.40 & 15.20 & 26.50 \\
& w/o Manual Fine-tune (Auto $\rightarrow$ Semi)                          & 80.30 & 65.50 & 52.70 & 14.30 & 27.80 \\
& Manual Only                                                             & 75.80 & 57.10 & 39.80 & 19.10 & 20.20 \\
\bottomrule
\end{tabular}
}
% 5. 终极空间压缩：如果前后正文间距还是太大，可以取消下一行的注释
% \vspace{-0.3cm} 
\end{table*}

\subsubsection{Robustness to Upstream Bounding-Box Errors}
\label{app:box_robustness}

DDVR-Net focuses on character recognition and restoration conditioned
on localized character regions, rather than jointly learning text
detection. The main evaluation therefore uses annotated character
boxes to isolate the proposed degradation-aware components. To
simulate errors introduced by an upstream detector, we perturb each
box independently by adding uniformly sampled offsets to its four
coordinates. The maximum offset is set to $5\%$, $10\%$, or $20\%$ of
the corresponding box width or height. We further simulate missed
detections by randomly removing $5\%$ or $10\%$ of the character
proposals. All variants use the same trained DDVR-Net without
additional fine-tuning.

\subsection{Case Study}
\label{app:case_study}
\begin{figure*}[h] % 如果是双栏论文想跨两栏显示，用 figure*；单栏用 figure
  \centering
    \includegraphics[width=1\linewidth]{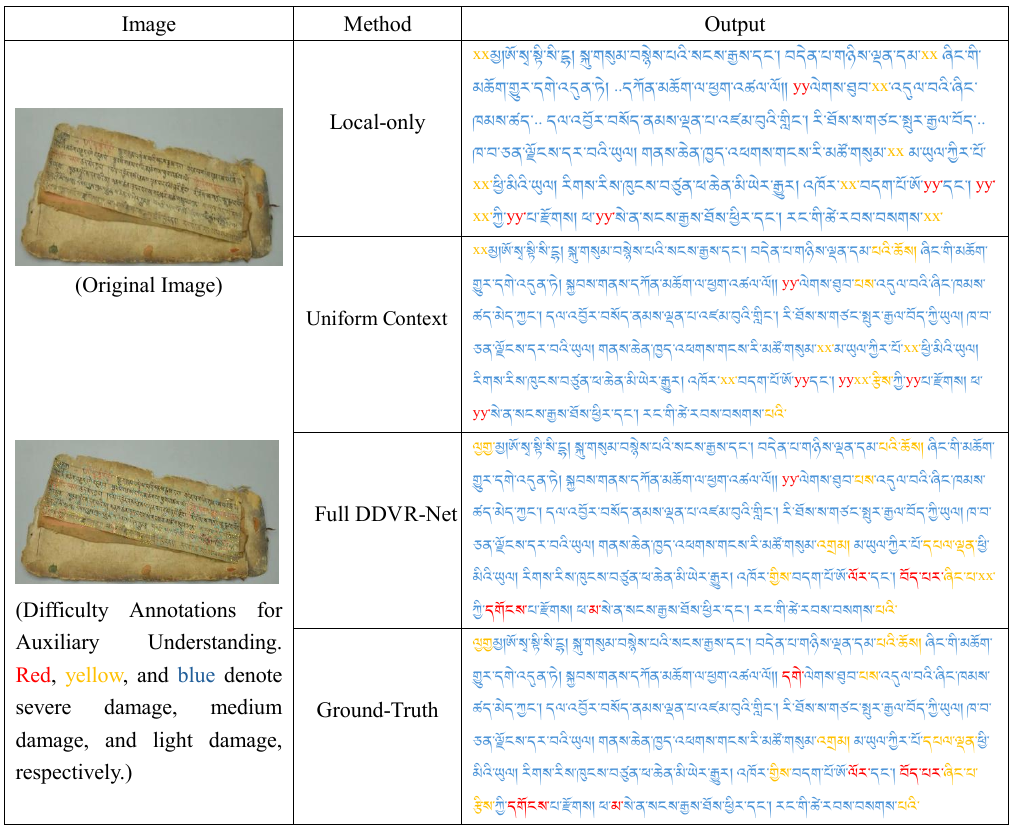} 
 \caption{Qualitative comparison on a mixed-visibility text line. To intuitively illustrate error locations,  recognition errors and hallucinations in lightly and moderately damaged regions are denoted by ‘\textcolor{blue}{xx}’ and ‘\textcolor{yellow}{yy}’, respectively, compared to the Ground-Truth. Our DDVR-Net successfully handles varying degradations.}
  \label{fig:case_study}
\end{figure*}

To intuitively demonstrate the efficacy of our visibility-guided fallback mechanism, Figure \ref{fig:case_study} presents a qualitative comparison of different restoration strategies on a severely degraded historical document page containing mixed-visibility characters. As highlighted in the comparison, for visually unambiguous characters (e.g., the blue bounding box), the \textit{Local-only} baseline and our DDVR-Net successfully recognize the target. In contrast, the \textit{Uniform Context} model suffers from hallucination and over-correction, overriding clear visual evidence with an incorrect language-driven guess. Conversely, for severely degraded regions where glyph structures are almost entirely destroyed (e.g., the red bounding box), the \textit{Local-only} baseline fails completely. Our DDVR-Net, dynamically triggered by the estimated visibility state, accurately recovers the missing semantics via the contextual decoder, successfully matching the restoration capability of the \textit{Uniform Context} approach. This visual evidence corroborates our quantitative findings: DDVR-Net elegantly resolves the mixed-visibility dilemma, achieving highly reliable restoration without compromising the integrity of clear historical text.

\bibliographystyle{ACM-Reference-Format}
\bibliography{sample-base}


\begin{thebibliography}{35}

%%% the Web address.

\ifx \showCODEN    \undefined \def \showCODEN     #1{\unskip}     \fi
\ifx \showISBNx    \undefined \def \showISBNx     #1{\unskip}     \fi
\ifx \showISBNxiii \undefined \def \showISBNxiii  #1{\unskip}     \fi
\ifx \showISSN     \undefined \def \showISSN      #1{\unskip}     \fi
\ifx \showLCCN     \undefined \def \showLCCN      #1{\unskip}     \fi
\ifx \shownote     \undefined \def \shownote      #1{#1}          \fi
\ifx \showarticletitle \undefined \def \showarticletitle #1{#1}   \fi
\ifx \showURL      \undefined \def \showURL       {\relax}        \fi
% invisible to TeX
\providecommand\bibfield[2]{#2}
\providecommand\bibinfo[2]{#2}
\providecommand\natexlab[1]{#1}
\providecommand\showeprint[2][]{arXiv:#2}

\bibitem[Achiam et~al\mbox{.}(2023)]%
        {achiam2023gpt}
\bibfield{author}{\bibinfo{person}{Josh Achiam}, \bibinfo{person}{Steven
  Adler}, \bibinfo{person}{Sandhini Agarwal}, \bibinfo{person}{Lama Ahmad},
  \bibinfo{person}{Ilge Akkaya}, \bibinfo{person}{Florencia~Leoni Aleman},
  \bibinfo{person}{Diogo Almeida}, \bibinfo{person}{Janko Altenschmidt},
  \bibinfo{person}{Sam Altman}, \bibinfo{person}{Shyamal Anadkat},
  {et~al\mbox{.}}} \bibinfo{year}{2023}\natexlab{}.
\newblock \showarticletitle{Gpt-4 technical report}.
\newblock \bibinfo{journal}{\emph{arXiv preprint arXiv:2303.08774}}
  (\bibinfo{year}{2023}).
\newblock


\bibitem[Aioanei et~al\mbox{.}(2024)]%
        {aioanei2024deep}
\bibfield{author}{\bibinfo{person}{Andrei~C Aioanei}, \bibinfo{person}{Regine~R
  Hunziker-Rodewald}, \bibinfo{person}{Konstantin~M Klein}, {and}
  \bibinfo{person}{Dominik~L Michels}.} \bibinfo{year}{2024}\natexlab{}.
\newblock \showarticletitle{Deep Aramaic: Towards a synthetic data paradigm
  enabling machine learning in epigraphy}.
\newblock \bibinfo{journal}{\emph{Plos one}} \bibinfo{volume}{19},
  \bibinfo{number}{4} (\bibinfo{year}{2024}), \bibinfo{pages}{e0299297}.
\newblock


\bibitem[Assael et~al\mbox{.}(2025)]%
        {assael2025contextualizing}
\bibfield{author}{\bibinfo{person}{Yannis Assael}, \bibinfo{person}{Thea
  Sommerschield}, \bibinfo{person}{Alison Cooley}, \bibinfo{person}{Brendan
  Shillingford}, \bibinfo{person}{John Pavlopoulos}, \bibinfo{person}{Priyanka
  Suresh}, \bibinfo{person}{Bailey Herms}, \bibinfo{person}{Justin Grayston},
  \bibinfo{person}{Benjamin Maynard}, \bibinfo{person}{Nicholas Dietrich},
  {et~al\mbox{.}}} \bibinfo{year}{2025}\natexlab{}.
\newblock \showarticletitle{Contextualizing ancient texts with generative
  neural networks}.
\newblock \bibinfo{journal}{\emph{Nature}} \bibinfo{volume}{645},
  \bibinfo{number}{8079} (\bibinfo{year}{2025}), \bibinfo{pages}{141--147}.
\newblock


\bibitem[Assael et~al\mbox{.}(2022)]%
        {assael2022restoring}
\bibfield{author}{\bibinfo{person}{Yannis Assael}, \bibinfo{person}{Thea
  Sommerschield}, \bibinfo{person}{Brendan Shillingford},
  \bibinfo{person}{Mahyar Bordbar}, \bibinfo{person}{John Pavlopoulos},
  \bibinfo{person}{Marita Chatzipanagiotou}, \bibinfo{person}{Ion
  Androutsopoulos}, \bibinfo{person}{Jonathan Prag}, {and}
  \bibinfo{person}{Nando De~Freitas}.} \bibinfo{year}{2022}\natexlab{}.
\newblock \showarticletitle{Restoring and attributing ancient texts using deep
  neural networks}.
\newblock \bibinfo{journal}{\emph{Nature}} \bibinfo{volume}{603},
  \bibinfo{number}{7900} (\bibinfo{year}{2022}), \bibinfo{pages}{280--283}.
\newblock


\bibitem[Bai et~al\mbox{.}(2025)]%
        {bai2025qwen3}
\bibfield{author}{\bibinfo{person}{Shuai Bai}, \bibinfo{person}{Yuxuan Cai},
  \bibinfo{person}{Ruizhe Chen}, \bibinfo{person}{Keqin Chen},
  \bibinfo{person}{Xionghui Chen}, \bibinfo{person}{Zesen Cheng},
  \bibinfo{person}{Lianghao Deng}, \bibinfo{person}{Wei Ding},
  \bibinfo{person}{Chang Gao}, \bibinfo{person}{Chunjiang Ge}, {et~al\mbox{.}}}
  \bibinfo{year}{2025}\natexlab{}.
\newblock \showarticletitle{Qwen3-vl technical report}.
\newblock \bibinfo{journal}{\emph{arXiv preprint arXiv:2511.21631}}
  (\bibinfo{year}{2025}).
\newblock


\bibitem[Bi et~al\mbox{.}(2025)]%
        {bi2025llava}
\bibfield{author}{\bibinfo{person}{Jinhe Bi}, \bibinfo{person}{Yujun Wang},
  \bibinfo{person}{Haokun Chen}, \bibinfo{person}{Xun Xiao},
  \bibinfo{person}{Artur Hecker}, \bibinfo{person}{Volker Tresp}, {and}
  \bibinfo{person}{Yunpu Ma}.} \bibinfo{year}{2025}\natexlab{}.
\newblock \showarticletitle{LLaVA steering: Visual instruction tuning with 500x
  fewer parameters through modality linear representation-steering}. In
  \bibinfo{booktitle}{\emph{Proceedings of the 63rd Annual Meeting of the
  Association for Computational Linguistics (Volume 1: Long Papers)}}.
  \bibinfo{pages}{15230--15250}.
\newblock


\bibitem[Cui et~al\mbox{.}(2025)]%
        {cui2025paddleocr}
\bibfield{author}{\bibinfo{person}{Cheng Cui}, \bibinfo{person}{Ting Sun},
  \bibinfo{person}{Manhui Lin}, \bibinfo{person}{Tingquan Gao},
  \bibinfo{person}{Yubo Zhang}, \bibinfo{person}{Jiaxuan Liu},
  \bibinfo{person}{Xueqing Wang}, \bibinfo{person}{Zelun Zhang},
  \bibinfo{person}{Changda Zhou}, \bibinfo{person}{Hongen Liu},
  {et~al\mbox{.}}} \bibinfo{year}{2025}\natexlab{}.
\newblock \showarticletitle{Paddleocr 3.0 technical report}.
\newblock \bibinfo{journal}{\emph{arXiv preprint arXiv:2507.05595}}
  (\bibinfo{year}{2025}).
\newblock


\bibitem[Ding et~al\mbox{.}(2025)]%
        {ding2025hisdoc}
\bibfield{author}{\bibinfo{person}{Kai Ding}, \bibinfo{person}{Sheng Jian},
  {and} \bibinfo{person}{Lianwen Jin}.} \bibinfo{year}{2025}\natexlab{}.
\newblock \showarticletitle{HisDoc-DETR: Integrating Semantic Learning and
  Feature Fusion for Historical Document Layout Analysis}. In
  \bibinfo{booktitle}{\emph{International Conference on Document Analysis and
  Recognition}}. Springer, \bibinfo{pages}{218--237}.
\newblock


\bibitem[Dong et~al\mbox{.}(2026)]%
        {dong2026mineru}
\bibfield{author}{\bibinfo{person}{Hejun Dong}, \bibinfo{person}{Junbo Niu},
  \bibinfo{person}{Bin Wang}, \bibinfo{person}{Weijun Zeng},
  \bibinfo{person}{Wentao Zhang}, {and} \bibinfo{person}{Conghui He}.}
  \bibinfo{year}{2026}\natexlab{}.
\newblock \showarticletitle{MinerU-Diffusion: Rethinking Document OCR as
  Inverse Rendering via Diffusion Decoding}.
\newblock \bibinfo{journal}{\emph{arXiv preprint arXiv:2603.22458}}
  (\bibinfo{year}{2026}).
\newblock


\bibitem[Duan et~al\mbox{.}(2026)]%
        {duan2026glm}
\bibfield{author}{\bibinfo{person}{Shuaiqi Duan}, \bibinfo{person}{Yadong Xue},
  \bibinfo{person}{Weihan Wang}, \bibinfo{person}{Zhe Su},
  \bibinfo{person}{Huan Liu}, \bibinfo{person}{Sheng Yang},
  \bibinfo{person}{Guobing Gan}, \bibinfo{person}{Guo Wang},
  \bibinfo{person}{Zihan Wang}, \bibinfo{person}{Shengdong Yan},
  {et~al\mbox{.}}} \bibinfo{year}{2026}\natexlab{}.
\newblock \showarticletitle{GLM-OCR Technical Report}.
\newblock \bibinfo{journal}{\emph{arXiv preprint arXiv:2603.10910}}
  (\bibinfo{year}{2026}).
\newblock


\bibitem[Guan et~al\mbox{.}(2025)]%
        {guan2025prep}
\bibfield{author}{\bibinfo{person}{Shuhao Guan}, \bibinfo{person}{Moule Lin},
  \bibinfo{person}{Cheng Xu}, \bibinfo{person}{Xinyi Liu},
  \bibinfo{person}{Jinman Zhao}, \bibinfo{person}{Jiexin Fan},
  \bibinfo{person}{Qi Xu}, {and} \bibinfo{person}{Derek Greene}.}
  \bibinfo{year}{2025}\natexlab{}.
\newblock \showarticletitle{Prep-OCR: a complete pipeline for document image
  restoration and enhanced OCR accuracy}. In
  \bibinfo{booktitle}{\emph{Proceedings of the 63rd Annual Meeting of the
  Association for Computational Linguistics (Volume 1: Long Papers)}}.
  \bibinfo{pages}{15413--15425}.
\newblock


\bibitem[Ham et~al\mbox{.}(2025)]%
        {ham2025glyph}
\bibfield{author}{\bibinfo{person}{Seungjae Ham}, \bibinfo{person}{Yeokyoung
  Yoon}, \bibinfo{person}{Il-Min Kim}, \bibinfo{person}{Sangseok Yun},
  \bibinfo{person}{Jae-Mo Kang}, {et~al\mbox{.}}}
  \bibinfo{year}{2025}\natexlab{}.
\newblock \showarticletitle{GLYPH-SR: Can We Achieve Both High-Quality Image
  Super-Resolution and High-Fidelity Text Recovery via VLM-Guided Latent
  Diffusion Model?}
\newblock  (\bibinfo{year}{2025}).
\newblock


\bibitem[Hegghammer(2022)]%
        {hegghammer2022ocr}
\bibfield{author}{\bibinfo{person}{Thomas Hegghammer}.}
  \bibinfo{year}{2022}\natexlab{}.
\newblock \showarticletitle{OCR with Tesseract, Amazon Textract, and Google
  Document AI: a benchmarking experiment}.
\newblock \bibinfo{journal}{\emph{Journal of Computational Social Science}}
  \bibinfo{volume}{5}, \bibinfo{number}{1} (\bibinfo{year}{2022}),
  \bibinfo{pages}{861--882}.
\newblock


\bibitem[Hu et~al\mbox{.}(2022)]%
        {hu2022lora}
\bibfield{author}{\bibinfo{person}{Edward~J Hu}, \bibinfo{person}{Yelong Shen},
  \bibinfo{person}{Phillip Wallis}, \bibinfo{person}{Zeyuan Allen-Zhu},
  \bibinfo{person}{Yuanzhi Li}, \bibinfo{person}{Shean Wang},
  \bibinfo{person}{Liang Wang}, \bibinfo{person}{Weizhu Chen}, {et~al\mbox{.}}}
  \bibinfo{year}{2022}\natexlab{}.
\newblock \showarticletitle{Lora: Low-rank adaptation of large language
  models.}
\newblock \bibinfo{journal}{\emph{Iclr}} \bibinfo{volume}{1},
  \bibinfo{number}{2} (\bibinfo{year}{2022}), \bibinfo{pages}{3}.
\newblock


\bibitem[Kanerva et~al\mbox{.}(2025)]%
        {kanerva2025ocr}
\bibfield{author}{\bibinfo{person}{Jenna Kanerva}, \bibinfo{person}{Cassandra
  Ledins}, \bibinfo{person}{Siiri K{\"a}pyaho}, {and} \bibinfo{person}{Filip
  Ginter}.} \bibinfo{year}{2025}\natexlab{}.
\newblock \showarticletitle{OCR error post-correction with LLMs in historical
  documents: no free lunches}. In \bibinfo{booktitle}{\emph{Proceedings of the
  Third Workshop on Resources and Representations for Under-Resourced Languages
  and Domains (RESOURCEFUL-2025)}}. \bibinfo{pages}{38--47}.
\newblock


\bibitem[Li et~al\mbox{.}(2025)]%
        {li2025dotsocrmultilingualdocumentlayout}
\bibfield{author}{\bibinfo{person}{Yumeng Li}, \bibinfo{person}{Guang Yang},
  \bibinfo{person}{Hao Liu}, \bibinfo{person}{Bowen Wang}, {and}
  \bibinfo{person}{Colin Zhang}.} \bibinfo{year}{2025}\natexlab{}.
\newblock \showarticletitle{dots.ocr: Multilingual Document Layout Parsing in a
  Single Vision-Language Model}.
\newblock  (\bibinfo{year}{2025}).
\newblock
\showeprint[arxiv]{2512.02498}~[cs.CV]
\urldef\tempurl%
\url{https://arxiv.org/abs/2512.02498}
\showURL{%
\tempurl}


\bibitem[Loshchilov and Hutter(2017)]%
        {loshchilov2017decoupled}
\bibfield{author}{\bibinfo{person}{Ilya Loshchilov} {and}
  \bibinfo{person}{Frank Hutter}.} \bibinfo{year}{2017}\natexlab{}.
\newblock \showarticletitle{Decoupled weight decay regularization}.
\newblock \bibinfo{journal}{\emph{arXiv preprint arXiv:1711.05101}}
  (\bibinfo{year}{2017}).
\newblock


\bibitem[Mazumder et~al\mbox{.}(2023)]%
        {mazumder2023dataperf}
\bibfield{author}{\bibinfo{person}{Mark Mazumder}, \bibinfo{person}{Colby
  Banbury}, \bibinfo{person}{Xiaozhe Yao}, \bibinfo{person}{Bojan
  Karla{\v{s}}}, \bibinfo{person}{William Gaviria~Rojas},
  \bibinfo{person}{Sudnya Diamos}, \bibinfo{person}{Greg Diamos},
  \bibinfo{person}{Lynn He}, \bibinfo{person}{Alicia Parrish},
  \bibinfo{person}{Hannah~Rose Kirk}, {et~al\mbox{.}}}
  \bibinfo{year}{2023}\natexlab{}.
\newblock \showarticletitle{Dataperf: Benchmarks for data-centric ai
  development}.
\newblock \bibinfo{journal}{\emph{Advances in Neural Information Processing
  Systems}}  \bibinfo{volume}{36} (\bibinfo{year}{2023}),
  \bibinfo{pages}{5320--5347}.
\newblock


\bibitem[Meng et~al\mbox{.}(2019)]%
        {meng2019glyce}
\bibfield{author}{\bibinfo{person}{Yuxian Meng}, \bibinfo{person}{Wei Wu},
  \bibinfo{person}{Fei Wang}, \bibinfo{person}{Xiaoya Li},
  \bibinfo{person}{Ping Nie}, \bibinfo{person}{Fan Yin}, \bibinfo{person}{Muyu
  Li}, \bibinfo{person}{Qinghong Han}, \bibinfo{person}{Xiaofei Sun}, {and}
  \bibinfo{person}{Jiwei Li}.} \bibinfo{year}{2019}\natexlab{}.
\newblock \showarticletitle{Glyce: Glyph-vectors for chinese character
  representations}.
\newblock \bibinfo{journal}{\emph{Advances in Neural Information Processing
  Systems}}  \bibinfo{volume}{32} (\bibinfo{year}{2019}).
\newblock


\bibitem[Nacar et~al\mbox{.}(2025)]%
        {nacar2025sard}
\bibfield{author}{\bibinfo{person}{Omer Nacar}, \bibinfo{person}{Yasser
  Al-Habashi}, \bibinfo{person}{Serry Sibaee}, \bibinfo{person}{Adel Ammar},
  {and} \bibinfo{person}{Wadii Boulila}.} \bibinfo{year}{2025}\natexlab{}.
\newblock \showarticletitle{SARD: A Large-Scale Synthetic Arabic OCR Dataset
  for Book-Style Text Recognition}.
\newblock \bibinfo{journal}{\emph{arXiv preprint arXiv:2505.24600}}
  (\bibinfo{year}{2025}).
\newblock


\bibitem[Nikolaidou et~al\mbox{.}(2022)]%
        {nikolaidou2022survey}
\bibfield{author}{\bibinfo{person}{Konstantina Nikolaidou},
  \bibinfo{person}{Mathias Seuret}, \bibinfo{person}{Hamam Mokayed}, {and}
  \bibinfo{person}{Marcus Liwicki}.} \bibinfo{year}{2022}\natexlab{}.
\newblock \showarticletitle{A survey of historical document image datasets}.
\newblock \bibinfo{journal}{\emph{International Journal on Document Analysis
  and Recognition (IJDAR)}} \bibinfo{volume}{25}, \bibinfo{number}{4}
  (\bibinfo{year}{2022}), \bibinfo{pages}{305--338}.
\newblock


\bibitem[Pardo et~al\mbox{.}(2024)]%
        {pardo2024advanced}
\bibfield{author}{\bibinfo{person}{Luc{\'\i}a~Pereira Pardo},
  \bibinfo{person}{Paul Dryburgh}, \bibinfo{person}{Elizabeth Biggs},
  \bibinfo{person}{Marc Vermeulen}, \bibinfo{person}{Peter Crooks},
  \bibinfo{person}{Adam Gibson}, \bibinfo{person}{Molly Fort},
  \bibinfo{person}{Constantina Vlachou-Mogire}, \bibinfo{person}{Moira
  Bertasa}, \bibinfo{person}{John~R Gilchrist}, {et~al\mbox{.}}}
  \bibinfo{year}{2024}\natexlab{}.
\newblock \showarticletitle{Advanced imaging to recover illegible text in
  historic documents. The challenge of past chemical treatments for ink
  enhancement}.
\newblock \bibinfo{journal}{\emph{Journal of Cultural Heritage}}
  \bibinfo{volume}{68} (\bibinfo{year}{2024}), \bibinfo{pages}{342--353}.
\newblock


\bibitem[Podell et~al\mbox{.}(2024)]%
        {podell2024sdxl}
\bibfield{author}{\bibinfo{person}{Dustin Podell}, \bibinfo{person}{Zion
  English}, \bibinfo{person}{Kyle Lacey}, \bibinfo{person}{Andreas Blattmann},
  \bibinfo{person}{Tim Dockhorn}, \bibinfo{person}{Jonas M{\"u}ller},
  \bibinfo{person}{Joe Penna}, {and} \bibinfo{person}{Robin Rombach}.}
  \bibinfo{year}{2024}\natexlab{}.
\newblock \showarticletitle{Sdxl: Improving latent diffusion models for
  high-resolution image synthesis}. In \bibinfo{booktitle}{\emph{International
  Conference on Learning Representations}}, Vol.~\bibinfo{volume}{2024}.
  \bibinfo{pages}{1862--1874}.
\newblock


\bibitem[Prasetiadi and Saputra(2024)]%
        {prasetiadi2024fragmented}
\bibfield{author}{\bibinfo{person}{Agi Prasetiadi} {and}
  \bibinfo{person}{Julian Saputra}.} \bibinfo{year}{2024}\natexlab{}.
\newblock \showarticletitle{Fragmented-cuneiform-based convolutional neural
  network for cuneiform character recognition}.
\newblock \bibinfo{journal}{\emph{IAES International Journal of Artificial
  Intelligence}} \bibinfo{volume}{13}, \bibinfo{number}{1}
  (\bibinfo{year}{2024}), \bibinfo{pages}{554}.
\newblock


\bibitem[Sadeghian et~al\mbox{.}(2024)]%
        {sadeghian2024cvocr}
\bibfield{author}{\bibinfo{person}{Rasoul Sadeghian}, \bibinfo{person}{Shahrooz
  Shahin}, {and} \bibinfo{person}{Sina Sareh}.}
  \bibinfo{year}{2024}\natexlab{}.
\newblock \showarticletitle{CVOCR: Context Vision OCR}. In
  \bibinfo{booktitle}{\emph{2024 20th IEEE/ASME International Conference on
  Mechatronic and Embedded Systems and Applications (MESA)}}. IEEE,
  \bibinfo{pages}{1--7}.
\newblock


\bibitem[Shi et~al\mbox{.}(2025)]%
        {shi2025large}
\bibfield{author}{\bibinfo{person}{Yongxin Shi}, \bibinfo{person}{Dezhi Peng},
  \bibinfo{person}{Yuyi Zhang}, \bibinfo{person}{Jiahuan Cao}, {and}
  \bibinfo{person}{Lianwen Jin}.} \bibinfo{year}{2025}\natexlab{}.
\newblock \showarticletitle{A large-scale dataset for Chinese historical
  document recognition and analysis}.
\newblock \bibinfo{journal}{\emph{Scientific Data}} \bibinfo{volume}{12},
  \bibinfo{number}{1} (\bibinfo{year}{2025}), \bibinfo{pages}{169}.
\newblock


\bibitem[{The Gemini Team}(2026)]%
        {team2024gemini}
\bibfield{author}{\bibinfo{person}{{The Gemini Team}}.}
  \bibinfo{year}{2026}\natexlab{}.
\newblock \bibinfo{title}{Gemini 3.1 Pro: A Smarter Model for Your Most Complex
  Tasks}.
\newblock \bibinfo{howpublished}{Google Blog}.
\newblock
\shownote{Published February 19, 2026}.
\newblock
\urldef\tempurl%
\url{https://blog.google/innovation-and-ai/models-and-research/gemini-models/gemini-3-1-pro/}
\showURL{%
\tempurl}


\bibitem[Wei et~al\mbox{.}(2025)]%
        {wei2025deepseek}
\bibfield{author}{\bibinfo{person}{Haoran Wei}, \bibinfo{person}{Yaofeng Sun},
  {and} \bibinfo{person}{Yukun Li}.} \bibinfo{year}{2025}\natexlab{}.
\newblock \showarticletitle{Deepseek-ocr: Contexts optical compression}.
\newblock \bibinfo{journal}{\emph{arXiv preprint arXiv:2510.18234}}
  (\bibinfo{year}{2025}).
\newblock


\bibitem[Wei et~al\mbox{.}(2026)]%
        {wei2026deepseek}
\bibfield{author}{\bibinfo{person}{Haoran Wei}, \bibinfo{person}{Yaofeng Sun},
  {and} \bibinfo{person}{Yukun Li}.} \bibinfo{year}{2026}\natexlab{}.
\newblock \showarticletitle{DeepSeek-OCR 2: Visual Causal Flow}.
\newblock \bibinfo{journal}{\emph{arXiv preprint arXiv:2601.20552}}
  (\bibinfo{year}{2026}).
\newblock


\bibitem[Yang et~al\mbox{.}(2025)]%
        {yang2025predicting}
\bibfield{author}{\bibinfo{person}{Zhenhua Yang}, \bibinfo{person}{Dezhi Peng},
  \bibinfo{person}{Yongxin Shi}, \bibinfo{person}{Yuyi Zhang},
  \bibinfo{person}{Chongyu Liu}, {and} \bibinfo{person}{Lianwen Jin}.}
  \bibinfo{year}{2025}\natexlab{}.
\newblock \showarticletitle{Predicting the original appearance of damaged
  historical documents}. In \bibinfo{booktitle}{\emph{Proceedings of the AAAI
  Conference on Artificial Intelligence}}, Vol.~\bibinfo{volume}{39}.
  \bibinfo{pages}{9382--9390}.
\newblock


\bibitem[Yu et~al\mbox{.}(2025)]%
        {yu2025docthinker}
\bibfield{author}{\bibinfo{person}{Wenwen Yu}, \bibinfo{person}{Zhibo Yang},
  \bibinfo{person}{Yuliang Liu}, {and} \bibinfo{person}{Xiang Bai}.}
  \bibinfo{year}{2025}\natexlab{}.
\newblock \showarticletitle{Docthinker: Explainable multimodal large language
  models with rule-based reinforcement learning for document understanding}. In
  \bibinfo{booktitle}{\emph{Proceedings of the IEEE/CVF International
  Conference on Computer Vision}}. \bibinfo{pages}{837--847}.
\newblock


\bibitem[Zhang et~al\mbox{.}(2025)]%
        {zhang2025reviving}
\bibfield{author}{\bibinfo{person}{Yuyi Zhang}, \bibinfo{person}{Peirong
  Zhang}, \bibinfo{person}{Zhenhua Yang}, \bibinfo{person}{Pengyu Yan},
  \bibinfo{person}{Yongxin Shi}, \bibinfo{person}{Pengwei Liu},
  \bibinfo{person}{Fengjun Guo}, {and} \bibinfo{person}{Lianwen Jin}.}
  \bibinfo{year}{2025}\natexlab{}.
\newblock \showarticletitle{Reviving cultural heritage: A novel approach for
  comprehensive historical document restoration}. In
  \bibinfo{booktitle}{\emph{Proceedings of the 63rd Annual Meeting of the
  Association for Computational Linguistics (Volume 1: Long Papers)}}.
  \bibinfo{pages}{28876--28892}.
\newblock


\bibitem[Zhao et~al\mbox{.}(2025)]%
        {zhao2025uni}
\bibfield{author}{\bibinfo{person}{Fangmin Zhao}, \bibinfo{person}{Weichao
  Zeng}, \bibinfo{person}{Zhenhang Li}, \bibinfo{person}{Dongbao Yang},
  \bibinfo{person}{Binbin Li}, \bibinfo{person}{Xiaojun Bi}, {and}
  \bibinfo{person}{Yu Zhou}.} \bibinfo{year}{2025}\natexlab{}.
\newblock \showarticletitle{Uni-DocDiff: A Unified Document Restoration Model
  Based on Diffusion}. In \bibinfo{booktitle}{\emph{Proceedings of the 33rd ACM
  International Conference on Multimedia}}. \bibinfo{pages}{8204--8213}.
\newblock


\bibitem[Zheng et~al\mbox{.}(2026)]%
        {zheng2026multimodal}
\bibfield{author}{\bibinfo{person}{Handong Zheng}, \bibinfo{person}{Yumeng Li},
  \bibinfo{person}{Kaile Zhang}, \bibinfo{person}{Liang Xin},
  \bibinfo{person}{Guangwei Zhao}, \bibinfo{person}{Hao Liu},
  \bibinfo{person}{Jiayu Chen}, \bibinfo{person}{Jie Lou},
  \bibinfo{person}{Jiyu Qiu}, \bibinfo{person}{Qi Fu}, {et~al\mbox{.}}}
  \bibinfo{year}{2026}\natexlab{}.
\newblock \showarticletitle{Multimodal OCR: Parse Anything from Documents}.
\newblock \bibinfo{journal}{\emph{arXiv preprint arXiv:2603.13032}}
  (\bibinfo{year}{2026}).
\newblock


\bibitem[Zhu et~al\mbox{.}(2024)]%
        {zhu2024reproducing}
\bibfield{author}{\bibinfo{person}{Shipeng Zhu}, \bibinfo{person}{Hui Xue},
  \bibinfo{person}{Na Nie}, \bibinfo{person}{Chenjie Zhu},
  \bibinfo{person}{Haiyue Liu}, {and} \bibinfo{person}{Pengfei Fang}.}
  \bibinfo{year}{2024}\natexlab{}.
\newblock \showarticletitle{Reproducing the past: a dataset for benchmarking
  inscription restoration}. In \bibinfo{booktitle}{\emph{Proceedings of the
  32nd ACM International Conference on Multimedia}}.
  \bibinfo{pages}{7714--7723}.
\newblock


\end{thebibliography}

\end{document}